\documentclass{article}

\usepackage{arxiv}

\usepackage[utf8]{inputenc}
\usepackage[T1]{fontenc}
\usepackage{hyperref}
\usepackage{url}
\usepackage{booktabs}
\usepackage{amsfonts}
\usepackage{nicefrac}
\usepackage[expansion=false]{microtype}
\usepackage{graphicx}
\usepackage{float}
\usepackage{amsmath,amssymb,amsthm}
\usepackage{multirow}
\usepackage{array}
\usepackage{xcolor}
\usepackage{caption}
\usepackage{enumitem}
\usepackage{subcaption}
\usepackage{cleveref}
\usepackage{setspace}
\usepackage{natbib}
\usepackage{doi}
\usepackage{placeins}
\usepackage{adjustbox}
\usepackage{pifont}
\usepackage{tabularx}
\usepackage{makecell}
\usepackage{ragged2e}
\usepackage{threeparttable}
\usepackage{xspace}
\usepackage{mathrsfs}
\usepackage{bm}

\usepackage{algorithm}
\usepackage{algpseudocode}

\usepackage{siunitx}
\DeclareSIUnit{\fps}{fps}
\DeclareSIUnit{\px}{px}
\DeclareSIUnit{\day}{d}

\newcolumntype{Y}{>{\RaggedRight\arraybackslash}X}
\graphicspath{{figures/}}

\title{Curvature-aware 3D length estimation of greenhouse cucumbers using RGB-D imaging and cubic spline arc-length integration}

\author{
  Manveen~Kaur \\
  Department of Mechanical Engineering\\
  University of Windsor\\
  Windsor, ON N9B 3P4, Canada \\
  \texttt{Kaur79@uwindsor.ca}
  \And
  Rajmeet~Singh$^{*}$ \\
Mohamed bin Zayed University of Artificial Intelligence\\
  Abu Dhabi, UAE \\
  \texttt{rajmeet.bhourji@mbzuai.ac.ae}
  \And
  Saeed~Mozaffari \\
  Department of Mechanical Engineering\\
  University of Windsor\\
  Windsor, ON N9B 3P4, Canada \\
  \texttt{s.mozaffari@uwindsor.ca}
  \And
  Majid~Ahmadi \\
  Department of Electrical and Computer Engineering\\
  University of Windsor\\
  Windsor, ON N9B 3P4, Canada \\
  \texttt{m.ahmadi@uwindsor.ca}
  \And
  Shahpour~Alirezaee \\
  Department of Mechanical Engineering\\
  University of Windsor\\
  Windsor, ON N9B 3P4, Canada \\
  \texttt{s.alirezaee@uwindsor.ca}
  \And
  \vspace{-1em}
  \normalsize\textit{$^{*}$ Corresponding author.}
}

\renewcommand{\shorttitle}{Curvature-aware 3D length estimation of greenhouse cucumbers}

\hypersetup{
  pdftitle={Curvature-aware 3D length estimation of greenhouse cucumbers using RGB-D imaging and cubic spline arc-length integration},
  pdfsubject={Computer Vision, Precision Agriculture, RGB-D Imaging, Cucumber Measurement},
  pdfauthor={Manveen Kaur, Rajmeet Singh, Saeed Mozaffari, Shahpour Alirezaee},
  pdfkeywords={Greenhouse cucumber, RGB-D imaging, Instance segmentation, Segment Anything Model, 3D length estimation, Cubic spline, Robotic harvesting, Multi-robot simulation},
  colorlinks=true,
  citecolor=blue,
  linkcolor=blue,
  urlcolor=blue
}

\begin{document}
\maketitle

\begin{abstract}
Greenhouse cucumbers are graded primarily by length, but manual
measurement is throughput-limited. This paper presents CucumberVision, a
non-contact RGB-D framework in which a YOLO26n instance-segmentation model
locates each fruit and the Segment Anything Model (SAM, ViT-B) refines its
mask. Five estimators are compared under matched conditions: a scan-line
baseline, two PCA variants, a keypoint-guided hybrid, and a proposed medial
arc spline (M5), which fits a cubic spline through 3D cross-section centroids
and integrates its arc length. The benchmark contains approximately 500
captures of 100 cucumbers spanning 6--40~cm with thread-based ground truth.
M5 gives the lowest error (MAE 0.58~cm; MAPE 4.13\%) and is the only estimator
without significant bias in the canopy benchmark. The ranking is replicated
under a second capture condition. Specimen-clustered bootstrap inference shows
M5 significantly outperforms the two non-SAM baselines, although its advantage
over the SAM-based methods is inconclusive. A controlled test also quantifies
approximately 33\% over-estimation when depth-stream rather than colour-stream
intrinsics are used after alignment. Finally, M5 is embedded in a simulated
three-robot harvesting fleet, where it grades all 133 harvested fruit into
length bins without extra viewpoints or coordination waits.
\end{abstract}


\section{Introduction}
\label{sec:introduction}

Global cucumber (\textit{Cucumis sativus} L.) production exceeded 88 million tonnes in 2023, making it the world's fourth most-produced vegetable \citep{fao2023}. In temperate regions, commercial production is predominantly conducted under protected cultivation. Cucumber quality is graded primarily by fruit length: European greenhouse standards commonly recognise Class~I (25--35~cm), Class~II (20--25~cm) and out-of-grade, with per-kilogram prices differing by 20--40\% between classes. Because cucumbers elongate at approximately 2--4~cm/d under optimal conditions \citep{neupane2021agronomy}, the number of fruit reaching a target length class within one to three days can be forecast from their current length distribution, enabling growers to schedule labour, pre-allocate packaging, and commit to distribution partners in advance.

The conventional measurement workflow, a worker with a digital caliper measuring each fruit individually achieves high accuracy but is throughput-limited at 300--400 fruit per hour and very tedious work. Real wages for hired farm workers have risen at 1.9\% per year over the past decade, roughly twice the long-run average rate, with the average nominal wage for non-supervisory crop and livestock workers reaching USD~18.13 per hour in 2024 \citep{usda2025}, intensifying pressure to automate inspection tasks. Consumer-grade depth cameras now achieve sub-centimetre accuracy at 0.3--1.5~m, integrate compactly with mobile robots, and retail for under USD~300 \citep{rijal2024}. However, camera-based length estimation accuracy is limited not by depth precision alone but by the algorithm used to identify the fruit's length axis. Five principal algorithmic families have emerged, four established and one proposed in this work:

\begin{itemize}
    \item \textbf{Dominant-axis skeleton sampling.} Pixels sampled along the longer bounding-box axis are deprojected to 3D \citep{wang2014, neupane2021agronomy}. Simple and fast but systematically underestimates diagonally oriented fruit due to cosine projection error.
    \item \textbf{Point-cloud PCA.} All depth-valid pixels are deprojected and the principal axis identified via SVD \citep{zheng2022, patel2025}. Orientation-invariant but susceptible to background leakage.
    \item \textbf{Foundation-model segmentation with mask-restricted principal-axis estimation.} SAM \citep{kirillov2023} yields a pixel-precise mask whose principal axis estimates the longitudinal axis \citep{zhou2023}. Eliminates background leakage but still measures a straight-line chord.
    \item \textbf{Semantic keypoint regression.} A pose network predicts anatomical landmarks deprojected to 3D and connected by piecewise segments \citep{zheng2022, ren2024}. Accurate under occlusion but requires keypoint annotations and underestimates arc length.
    \item \textbf{3D medial arc spline fitting (proposed).} The SAM mask is deprojected, cross-section centroids along the principal axis are computed, and a chord-length-parameterised cubic spline is fitted. Arc length is obtained by numerical integration, providing continuous curvature-aware measurement. Curve-based length recovery has been applied to fish \citep{zhou2023} and other elongated organisms; the contribution here is its combination with foundation-model segmentation for greenhouse fruit sizing.
\end{itemize}

No published study has evaluated all five families under matched conditions. This paper addresses this gap through five research contributions and one system component:

\begin{itemize}
    \item \textbf{Unified five-method pipeline.} CucumberVision implements M1--M5 sharing a single RGB-D camera, YOLO26n detector, SAM ViT-B segmentor, burst depth averaging, and colour-stream intrinsic alignment.
    \item \textbf{Novel curvature-aware method.} M5 deprojects SAM-masked pixels into 3D, fits a cubic spline through principal-axis cross-section centroids, and integrates arc length.
    \item \textbf{Controlled comparative evaluation.} All five methods benchmarked across three size categories with specimen-clustered bootstrap confidence intervals that respect the non-independence of repeated captures of the same fruit.
    \item \textbf{System component---adaptive method selector.} A cascading decision tree assigns the most accurate available method per fruit and guarantees 100\% measurement coverage. This is reported as an engineering component of the deployed system rather than as a research contribution: on the present benchmark the primary branch (M5) was selected for every capture, so the fallback paths were not exercised.
    \item \textbf{Deployable dashboard.} CucumberVision Streamlit app with live camera and upload modes, per-fruit visualisation, and CSV/Excel export is developed.
    \item \textbf{Quantified intrinsics-alignment error.} A controlled before/after test on the benchmark captures shows that deprojecting with depth-stream intrinsics after \texttt{rs.align(rs.stream.color)} over-estimates length by approximately 33\%, an error source that appears under-reported in the depth-camera measurement literature.
    \item \textbf{System demonstration---simulated robotic harvest-and-grade loop.} The M5 measurement chain is embedded in the AgriBot Robotics Simulator, a browser-based greenhouse in which three mobile manipulators (two Clearpath Husky A200 and one Omron LD-90, each carrying a Franka arm with a wrist-mounted RealSense D435i) harvest cucumbers and sort them into length bins. This shows how per-fruit arc length can drive a grading decision inside a multi-robot harvesting workflow. It is reported as a system demonstration rather than as accuracy evidence, because the simulator emulates the detector and spline from known fruit geometry.
\end{itemize}

The remainder is organised as follows. Section~\ref{sec:related_work} reviews related work. Section~\ref{sec:methodology} presents the methodology. Section~\ref{sec:experimentation} describes the experimental protocol. Section~\ref{sec:results} presents results and deployment. Section~\ref{sec:simulation} integrates M5 into a simulated multi-robot harvesting and grading fleet. Section~\ref{sec:discussion} discusses findings and limitations. Section~\ref{sec:conclusions} concludes.

\section{Related Work}
\label{sec:related_work}

\subsection{Non-contact fruit and vegetable sizing}
\citet{wang2014} estimated sweet onion diameter from a Kinect~v1 using bounding-box projection, forming the conceptual basis of M1. \citet{neupane2021agronomy} compared four depth sensors for orchard sizing, concluding that the D435i offered the best accuracy below 1.5~m. \citet{liu2019} applied Mask R-CNN to greenhouse cucumbers with stereo depth, reporting 10--15\% length errors. \citet{lawal2024} compared YOLO variants for cucurbit detection, motivating the use of YOLO26. \citet{neupane2022} compared three RGB-D sizing approaches for on-the-go mango measurement against caliper ground truth, obtaining a lowest RMSE of 4.7 and 5.1~mm on lineal dimensions for two varieties ($n = 104$); partly occluded fruit were excluded by design. \citet{song2023} reconstructed field crop canopies with a consumer RGB-D camera on a moving platform, recovering plant height ($R^2$ 0.90--0.96, RMSE 15--22~mm) and leaf area for seedling-stage corn, tobacco, potato and \textit{Bletilla striata}. \citet{zhou2023} measured curved fish bodies by stereo vision with skeleton extraction and refraction correction, reporting 0.9\% mean relative error. \citet{hong2024} achieved sub-centimetre accuracy for oriental melon via multi-view fusion but required impractical infrastructure. \citet{turkseven2021} demonstrated D435 suitability for greenhouse tasks. \citet{singh2024, singh2025} deployed the D435 for tomato flower detection on robotic platforms.

\subsection{Deep-learning detectors and segmenters}
The YOLO family \citep{redmon2016} established single-stage detection at real-time speed. \citet{zhao2024} extended YOLOv7 with an instance-segmentation module and a keypoint mechanism for 6-DoF pose estimation in selective harvesting, validated on cucumber datasets. The Ultralytics framework \citep{ultralytics2024} provides the training backbone used here. \citet{liu2019} reported 85.3\% mAP for cucumber detection with Mask R-CNN \citep{he2017}, but its higher latency motivated the present use of YOLO26n with integrated segmentation.

\subsection{SAM in agricultural applications}
SAM \citep{kirillov2023} segments arbitrary objects given a prompt. \citet{carraro2023} identified SAM as transformative for smart farming. \citet{williams2024} demonstrated competitive zero-shot leaf segmentation. \citet{kaur2025} deployed a visual-language transformer for greenhouse disease detection, establishing the Streamlit dashboard paradigm adopted here. SAM~2 \citep{ravi2024} offers a path to multi-frame smoothing discussed in Section~\ref{sec:discussion}. In this work, SAM refines YOLO26 detections to pixel-precise masks for M3--M5.

\subsection{Keypoint detection for produce measurement}
\citet{zheng2022} located six keypoints per vegetable and estimated length as the top-to-bottom chord scaled by the depth of the object centre, evaluated on rigid plastic models under uniform laboratory illumination; best-case error was near 2\% (tomato diameter, eggplant length) at a 40~cm working distance, while cucumber diameter error exceeded 6\% under the same conditions. \citet{ren2024} extended the approach to irregular root vegetables. The present work implements a YOLO26-pose head with five keypoints per cucumber integrated into an adaptive selector.

\subsection{Medial-axis and spline-based measurement}
The medial-axis algorithm of \citet{lee1994} extracts topological centre-lines and has been applied to fish length \citep{zhou2023}, rice panicles \citep{wu2019}, wheat \citep{aich2017}, and maize ears \citep{bao2019}. However, 2D arc length suffers perspective distortion \citep{paulus2019}. Spline fitting provides rigorous 3D arc-length computation \citep{deboor2001, farin2002}: \citet{paulus2019} fitted B-splines to barley ears, \citet{jin2018} to maize stalks, and \citet{magistri2022} to crop rows. \citet{tagliasacchi2016} reviewed 3D skeleton extraction from point clouds. Curve-based length recovery for elongated organisms is established---skeleton extraction for curved fish \citep{zhou2023} and parametric curve modelling for sea cucumbers---but we are not aware of prior work combining foundation-model segmentation, principal-axis centroid binning and cubic spline arc-length integration for greenhouse fruit sizing. M5 addresses this gap.

\subsection{Research gap analysis}
Three gaps motivate this work.
\begin{itemize}
    \item \textbf{Gap~1:} No controlled multi-method comparison of all five algorithmic families on the same cucumber dataset.
    \item \textbf{Gap~2:} No published method computes 3D arc length of curved vegetables via spline fitting, despite a chord-to-arc bias of $(L - c)/L \approx L^2/(24R^2)$, which for the present specimens ($L = 6.9$--38.0~cm across the primary benchmark and held-out set, $R = 30$--60~cm) ranges from below 0.1\% to about 3\%.
    \item \textbf{Gap~3:} No adaptive selector dynamically assigns measurement methods per fruit based on real-time input availability.
\end{itemize}
Table~\ref{tab:related_work} summarises representative prior studies.

\begin{table}[t]
\centering
\caption{Representative related studies, with author lists, sensors and reported metrics verified against the published sources. Note that reported accuracy is not comparable across rows: the measurand and ground-truth (GT) instrument differ.}
\label{tab:related_work}
\scriptsize
\setlength{\tabcolsep}{3pt}
\renewcommand{\arraystretch}{1.08}
\begin{tabularx}{\linewidth}{@{}
  >{\hsize=.90\hsize\RaggedRight\arraybackslash}X
  >{\hsize=1.15\hsize\RaggedRight\arraybackslash}X
  >{\hsize=1.15\hsize\RaggedRight\arraybackslash}X
  >{\hsize=1.00\hsize\RaggedRight\arraybackslash}X
  >{\hsize=1.00\hsize\RaggedRight\arraybackslash}X
  >{\hsize=.80\hsize\RaggedRight\arraybackslash}X@{}}
\toprule
\textbf{Reference} & \textbf{Crop} & \textbf{Sensor} & \makecell[l]{\textbf{GT}\\\textbf{instrument}} & \textbf{Measurand} & \makecell[l]{\textbf{Reported}\\\textbf{accuracy}} \\ \midrule
\citet{wang2014} & Sweet onion & Kinect v1 & Caliper & Diameter (chord) & $\approx 8\%$ \\[3pt]
\citet{liu2019} & Cucumber & Stereo RGB & Manual & Length (chord) & $\approx 12\%$ \\[3pt]
\citet{zheng2022} & Cucumber, eggplant, tomato, pepper & RealSense D415, 640$\times$480 & Model specification & 2-keypoint chord & $\approx 2\%$ best case; cucumber diameter $>6\%$ \\[3pt]
\citet{neupane2022} & Mango & ToF RGB-D & Caliper & Lineal (chord) & RMSE 4.7\,/\,5.1\,mm \\[3pt]
\citet{zhou2023} & Fish & Stereo RGB-D & Manual & Length via skeleton & 0.9\% \\[3pt]
\citet{song2023} & Corn, tobacco, potato, Bletilla & Consumer RGB-D & Manual & Plant height, leaf area & Height RMSE 15--22\,mm \\[3pt]
\citet{paulus2019} & Barley & Multi-view & Manual & B-spline on 3D cloud & --- \\[3pt]
\citet{hong2024} & Oriental melon & Multi RGB-D & Manual & 3D multi-view & $\sim 1.5\%$ \\[3pt]
\textbf{Our proposed work (2026)} & \textbf{Cucumber} & \textbf{RGB-D + YOLO26 + SAM} & \textbf{Thread} & \textbf{3D arc length} & \textbf{MAE 5.8\,mm, RMSE 8.9\,mm, MAPE 4.13\%} \\ \bottomrule
\end{tabularx}
\end{table}

\section{Methodology}
\label{sec:methodology}
\begin{figure}[!t]
\centering
\includegraphics[width=0.85\linewidth]{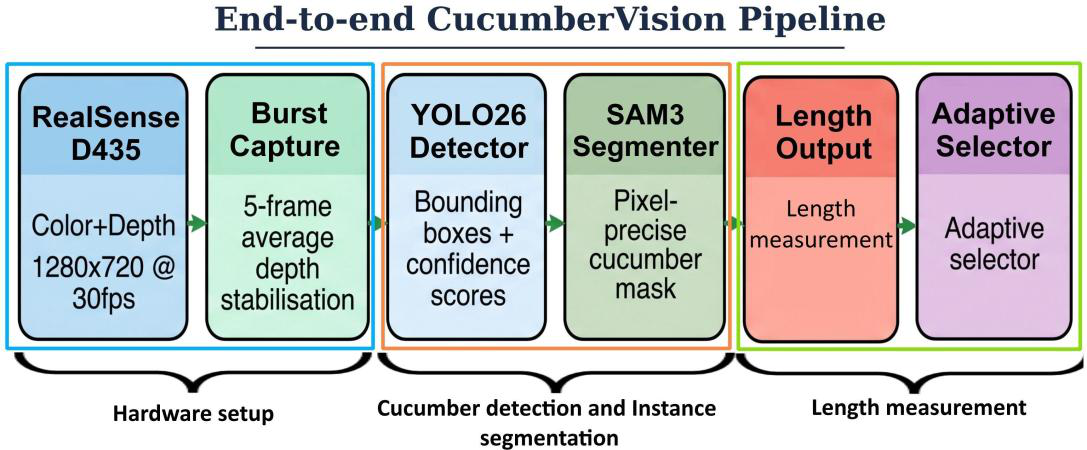}
\caption{End-to-end pipeline: D435 burst capture $\rightarrow$ YOLO26n detection $\rightarrow$ SAM mask refinement $\rightarrow$ adaptive method selection (M1--M5) $\rightarrow$ annotated length output.}
\label{fig:pipeline}
\end{figure}

This section describes the CucumberVision pipeline from sensor acquisition through detection, segmentation, and the five measurement methods. Fig.~\ref{fig:pipeline} provides an overview.

\subsection{Hardware configuration}
The Intel RealSense D435 active-stereo RGB-D camera was deployed in two environments: a controlled laboratory (University of Windsor) and a commercial greenhouse (Prism farm, Canada). The camera was positioned 0.27--0.77~m from the canopy (logged working-distance range across all benchmark captures). Table~\ref{tab:camera_params} summarises key parameters.

At each position, a 5-frame burst was acquired. Depth frames were pixel-wise median-averaged ($\sim 44\%$ noise reduction); the sharpest colour frame was selected by Laplacian variance \citep{pech2000}.

\begin{table}[htbp]
\centering
\caption{Intel RealSense D435 camera parameters.}
\label{tab:camera_params}
\footnotesize
\begin{tabularx}{\linewidth}{lX}
\toprule
\textbf{Parameter} & \textbf{Value} \\ \midrule
Resolution (colour \& depth) & 1280 $\times$ 720 px @ 30 fps \\
Depth unit / operating range & 1 mm per count; 0.1--5.0 m \\
Focal length $f_x, f_y$ & 910.22, 910.39 px \\
Principal point $(c_x, c_y)$ & (642.8, 358.5) px \\
Burst capture & 5 frames; median depth; sharpest colour \\ \bottomrule
\end{tabularx}
\end{table}

Depth frames are aligned to the colour stream with \texttt{rs.align(rs.stream.color)} before any measurement. After this operation each depth sample is expressed on the colour-sensor pixel grid, so Eq.~(\ref{eq:deproject}) must be evaluated with colour-stream intrinsics. The D435 depth module has a wider field of view than its colour module ($87^{\circ}$ against $70.2^{\circ}$ as calibrated on our unit), and correspondingly a shorter focal length ($f_x = 674.4$~px against 910.2~px at 1280 $\times$ 720). Substituting the depth focal length into the denominator of Eq.~(\ref{eq:deproject}) therefore inflates the recovered $X$ and $Y$ coordinates by the ratio of the two focal lengths, 1.35.

To quantify the consequence we re-ran every measurement method on all benchmark captures, changing nothing but the intrinsics passed to the deprojection. Table~\ref{tab:intrinsics_error} reports the result. The error is a systematic over-estimation of roughly one third for every method. It is not a uniform rescaling: $f_x$ and $f_y$ scale $X$ and $Y$ but leave $Z$ unchanged, so the magnitude depends on each fruit's orientation relative to the optical axis, producing the spread shown in the final column. We note this explicitly because the \texttt{pyrealsense2} alignment call does not itself alter the intrinsics object a user is holding, so the substitution is silent and produces plausible-looking lengths.

\begin{table}[htbp]
\centering
\caption{Effect of deprojecting with depth-stream rather than colour-stream intrinsics after alignment to the colour stream, on the benchmark captures. Only the intrinsics differ between the two columns. Depth-stream intrinsics are the D435 nominal values for the $87^{\circ}$ depth field of view.}
\label{tab:intrinsics_error}
\scriptsize
\setlength{\tabcolsep}{4pt}
\begin{tabularx}{\linewidth}{@{}l c c c >{\centering\arraybackslash}X@{}}
\toprule
\textbf{Method} & \makecell{\textbf{MAE, colour}\\\textbf{(cm)}} & \makecell{\textbf{MAE, depth}\\\textbf{(cm)}} & \makecell{\textbf{Mean}\\\textbf{bias}} & \makecell{\textbf{Range of}\\\textbf{bias}} \\
\midrule
M1 & 1.674 & 3.323 & +33.7\% & +23.0\% to +35.0\% \\
M2 & 0.848 & 5.018 & +33.9\% & +18.6\% to +53.4\% \\
M3 & 0.858 & 4.404 & +34.5\% & +31.9\% to +36.8\% \\
M5 & 0.646 & 4.745 & +32.6\% & +26.3\% to +47.9\% \\
\bottomrule
\end{tabularx}
\end{table}

\subsection{Cucumber detection and instance segmentation}
\begin{figure}[!t]
\centering
\includegraphics[width=0.85\linewidth]{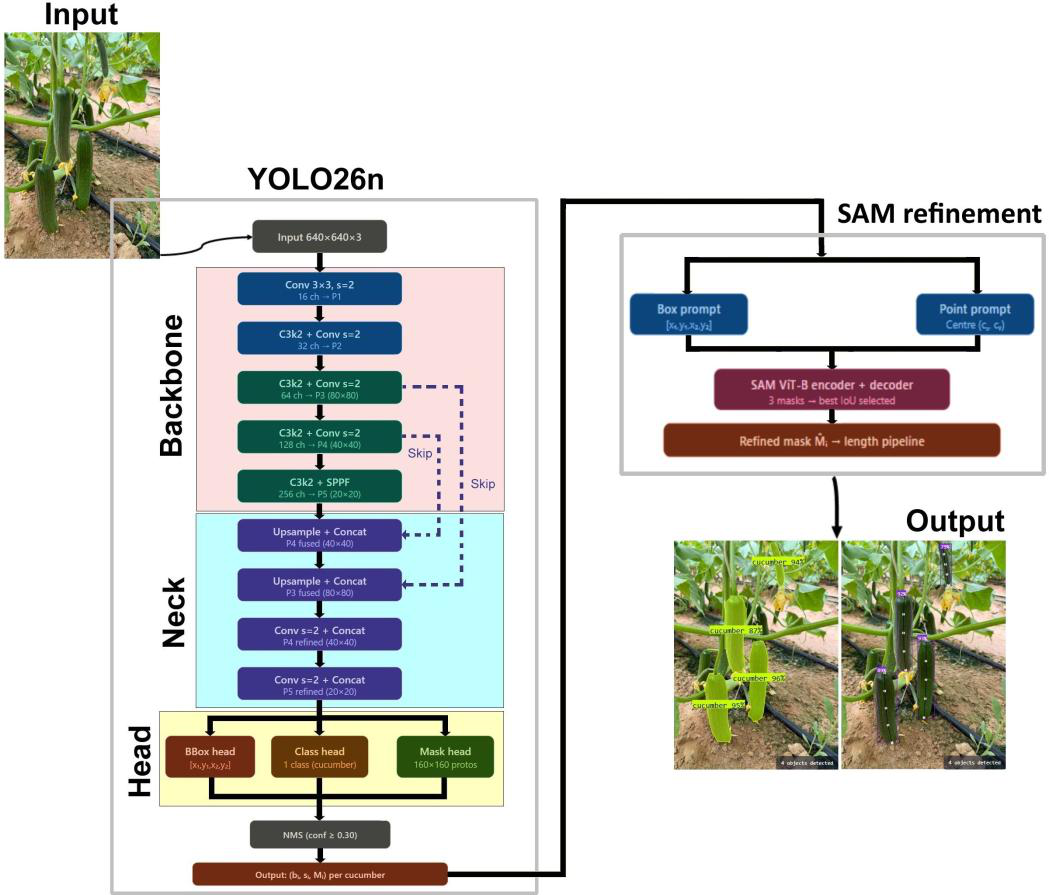}
\caption{Two-stage pipeline: YOLO26n backbone--neck--head with C3k2 blocks, FPN+PAN, and decoupled heads, followed by SAM ViT-B mask refinement via box and centre-point prompting. Representative input/output shown.}
\label{fig:yolo_sam}
\end{figure}

Fig.~\ref{fig:yolo_sam} presents the YOLO26n + SAM ViT-B two-stage pipeline. The YOLO26n backbone extracts features at P3 ($80 \times 80$), P4 ($40 \times 40$), and P5 ($20 \times 20$) using C3k2 blocks and SPPF. The FPN+PAN neck fuses features bidirectionally. Three decoupled heads predict bounding boxes, class confidence, and $160 \times 160$ prototype masks, filtered by NMS at confidence 0.30. Each detection is forwarded to SAM ViT-B (\texttt{sam\_vit\_b\_01ec64.pth}) using the bounding box as a box prompt and its centre as a foreground point prompt. With \texttt{multimask\_output=True}, the highest-IoU mask is selected. The ViT-B encoder runs once per frame, amortised across all detections. SAM succeeded on 98\% of instances; the 2\% failure rate corresponds to heavily occluded specimens.

The YOLO26 Instance Segmentation (Ultralytics (version 8.4+)) variant was selected because it balances accuracy, segmentation quality, and edge deployability within a single architecture. With only $\sim 2.4$~M parameters and $\sim 6.5$~GFLOPs, it executes in under 10~ms per frame on an NVIDIA RTX 3060, matching the D435's 30~fps and remaining viable on embedded platforms such as NVIDIA Jetson. The architecture uses C3k2 blocks with cross-stage partial connections and an SPPF module for multi-scale representation. Its native instance segmentation with $160 \times 160$ prototypes and decoupled head design stabilises gradient flow on small single-class datasets.

\subsection{Length measurement and adaptive method selector}
\begin{figure}[!t]
\centering
\includegraphics[width=0.95\linewidth]{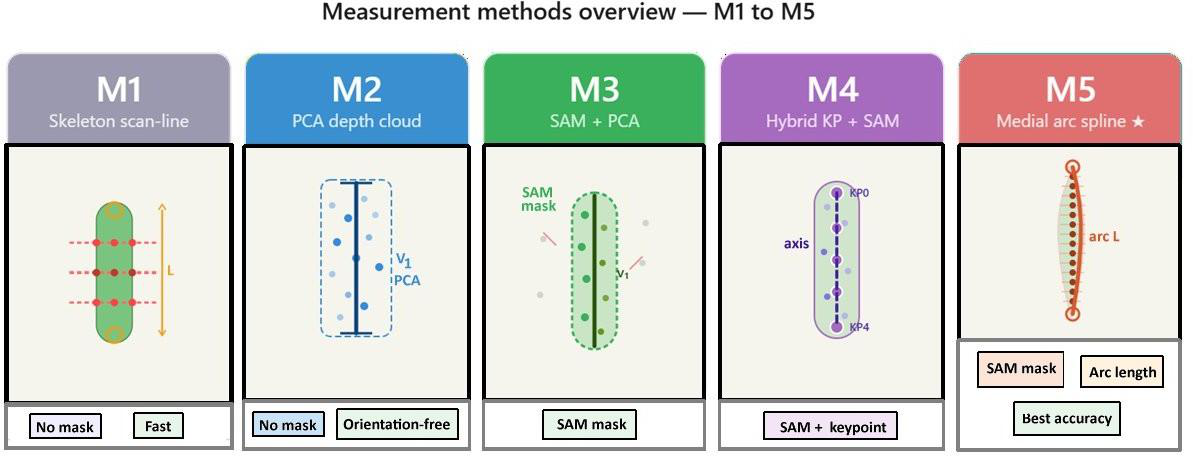}
\caption{Overview of M1--M5. M1: scan-line (no mask, fast). M2: PCA (orientation-free). M3: SAM + medial axis. M4: keypoints + SAM. M5: medial arc spline ($\star$, best accuracy).}
\label{fig:methods_overview}
\end{figure}

Each cucumber instance is represented by a refined mask $\hat{M}_i$, bounding box $b_i = [x_1, y_1, x_2, y_2]$, confidence $s_i$, and aligned depth map $D \in \mathbb{R}^{H \times W}$. Five methods address different trade-offs between cost, inputs, and accuracy. Fig.~\ref{fig:methods_overview} provides a visual overview.

\subsubsection{M1: Three-line skeleton scan baseline}
\begin{figure}[htbp]
\centering
\includegraphics[width=0.75\linewidth]{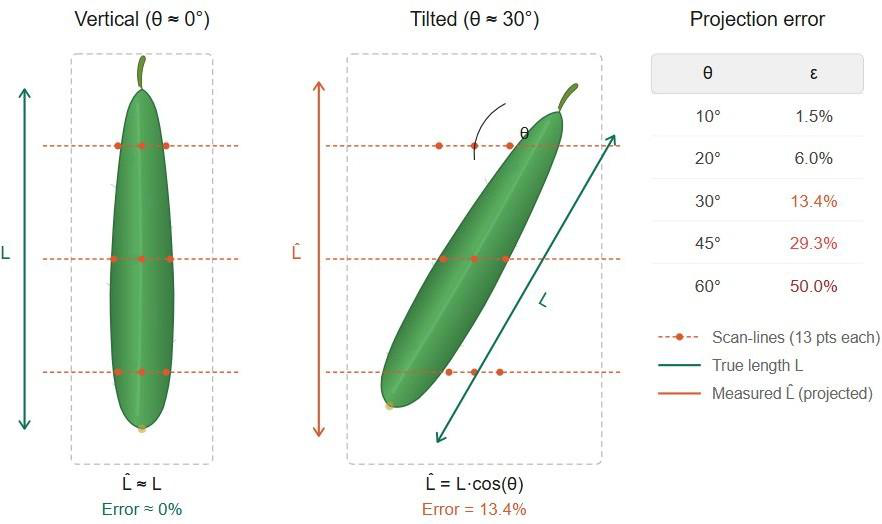}
\caption{M1 geometric limitation: vertical cucumber measured correctly (left); tilted cucumber underestimated by $1 - \cos(\theta)$ (centre); error table (right).}
\label{fig:m1_error}
\end{figure}

M1 casts three scan-lines along the dominant bounding-box axis, samples 13 depth points each ($11 \times 11$ median filter), discards invalid readings, deprojects the outermost valid pair to 3D, and returns their Euclidean distance ($\sim 28$~ms). As shown in Fig.~\ref{fig:m1_error}, for a fruit whose major axis makes an in-image-plane angle $\theta$ with the dominant bounding-box axis so that the outermost points sampled by the axis-aligned scan-lines are separated along the fruit by $L \cos\theta$ rather than by $L$:
\begin{equation}
\hat{L}_{\text{M1}} = L \cdot \cos(\theta), \quad \varepsilon_{\text{M1}} = 1 - \cos(\theta)
\label{eq:m1}
\end{equation}
yielding 13.4\% underestimation at $30^{\circ}$ and 29.3\% at $45^{\circ}$.

\subsubsection{M2: PCA on bounding-box depth point cloud}
\begin{figure}[htbp]
\centering
\includegraphics[width=0.75\linewidth]{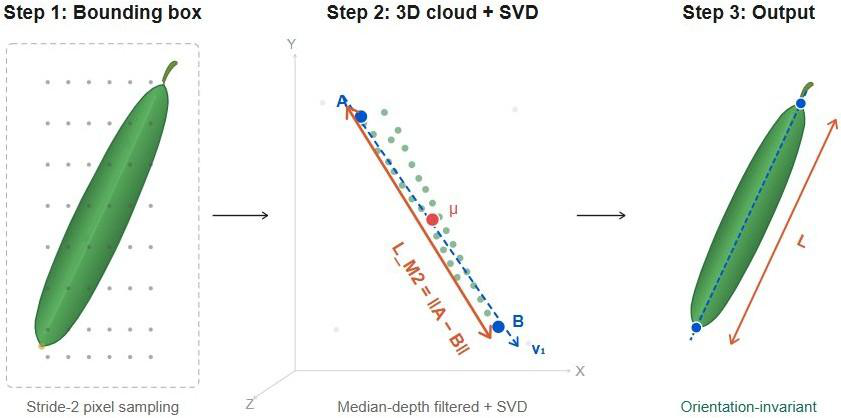}
\caption{M2: stride-2 sampling (Step 1), SVD principal axis with endpoints A, B (Step 2), orientation-invariant output (Step 3).}
\label{fig:m2_pca}
\end{figure}

M2 deprojects bounding-box pixels (stride 2) to 3D, filters by median depth ($|z_i - \tilde{z}| \le 0.035$~m), and applies SVD to find the principal axis $\mathbf{v}_1$ (Fig.~\ref{fig:m2_pca}):
\begin{equation}
\boldsymbol{\mu} = \frac{1}{N} \sum_i \mathbf{p}_i, \quad \tilde{\mathbf{P}} = \mathbf{P} - \boldsymbol{\mu} = \mathbf{U} \boldsymbol{\Sigma} \mathbf{V}^T, \quad \mathbf{v}_1 = \mathbf{V}^T[0, :]
\label{eq:pca}
\end{equation}
Endpoints are the extreme projections onto $\mathbf{v}_1$: $\hat{L}_{\text{M2}} = \|\mathbf{A} - \mathbf{B}\|$. M2 is invariant to in-image-plane orientation but susceptible to background leakage; the fixed 3.5~cm depth gate also truncates specimens oriented substantially along the optical axis, so it is not invariant to out-of-plane tilt.

\subsubsection{M3: SAM mask with PCA on masked point cloud}
\begin{figure}[htbp]
\centering
\includegraphics[width=0.75\linewidth]{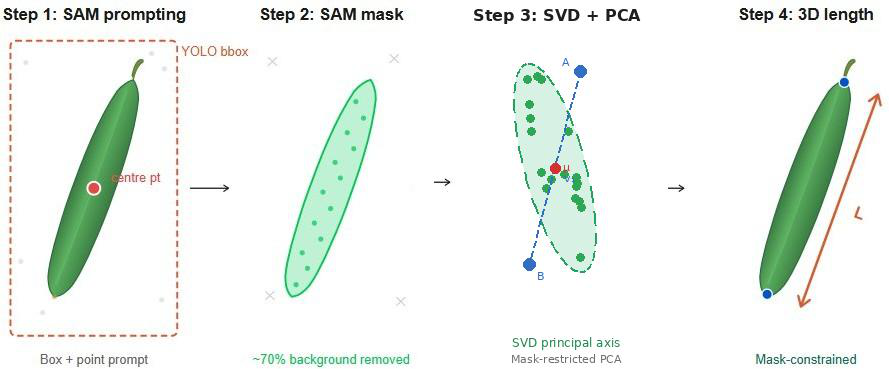}
\caption{M3: SAM prompting with box + centre-point (Step 1), mask generation---highest-IoU mask selected (Step 2), masked point cloud construction (Step 3), SVD/PCA principal axis and 3D endpoint extraction (Step 4).}
\label{fig:m3_sam_pca}
\end{figure}

M3 restricts the depth point cloud to SAM-masked pixels, eliminating approximately 70\% of background pixels that contaminate the M2 bounding-box cloud, as depicted in Fig.~\ref{fig:m3_sam_pca}. The masked cloud is then processed by the identical SVD/PCA procedure as M2 (Eq.~\ref{eq:pca}): centroid subtraction, thin SVD, and principal-axis extraction. The farthest projections onto $\mathbf{v}_1$ give 3D endpoints $\mathbf{A}, \mathbf{B}$ and $\hat{L}_{\text{M3}} = \|\mathbf{A} - \mathbf{B}\|$ ($\sim 65$~ms). M3 falls back to M2 if the SAM mask contains fewer than 50 valid depth pixels.

However, M3 inherits two limitations: SAM mask boundary imprecision on small cucumbers can cause the mask to bleed onto adjacent green background, introducing noisy pixels that pull the PCA axis off-centre; and the chord-based endpoint measurement still underestimates curved specimens (same as M2). These motivate M4, which replaces geometric axis estimation with semantically defined keypoints.

\subsubsection{M4: Hybrid keypoint-guided SAM cloud}
\begin{figure}[htbp]
\centering
\includegraphics[width=0.75\linewidth]{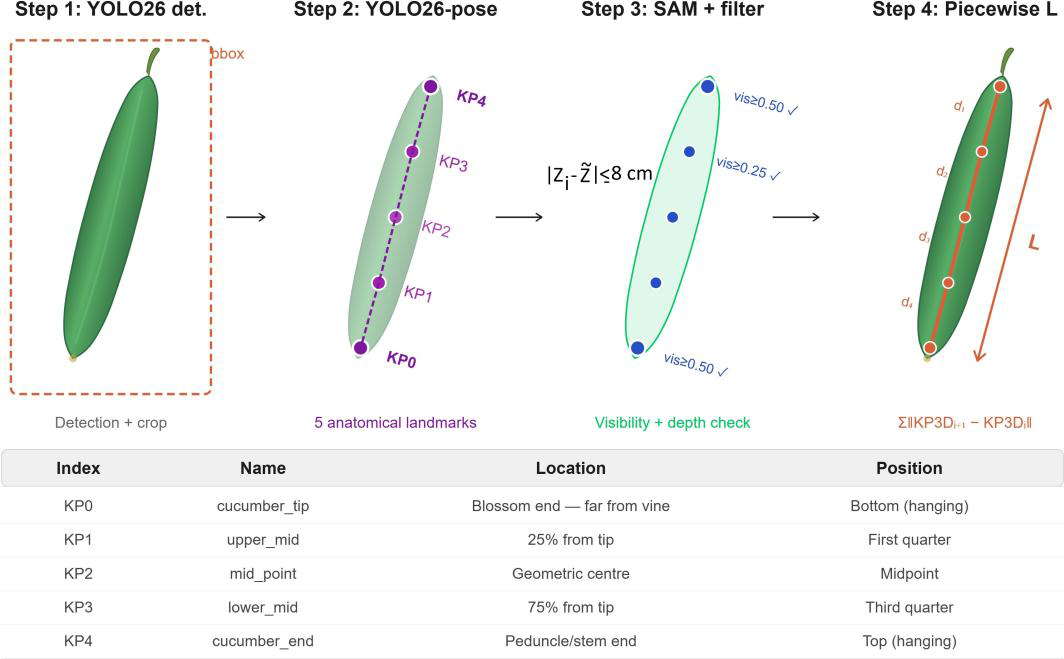}
\caption{M4: YOLO26n detection (Step 1), YOLO26-pose keypoints KP0--KP4 (Step 2), visibility and depth filtering (Step 3), piecewise 3D length (Step 4). Keypoint definitions tabulated.}
\label{fig:m4_keypoints}
\end{figure}

M4 uses a YOLO26-pose model to predict five anatomical keypoints (KP0--KP4) from blossom to stem end (Fig.~\ref{fig:m4_keypoints}). Endpoint keypoints KP0 and KP4 require visibility $v_i \ge 0.50$; intermediate keypoints KP1--KP3 accept $v_i \ge 0.25$. A depth filter rejects any keypoint whose depth deviates by more than 8~cm from the median keypoint depth, i.e.\ $|z_i - \text{median}(z_j)| > 0.08$~m. With $N_{\text{vis}} \ge 4$:
\begin{equation}
\hat{L}_{\text{M4}} = \sum_{i=0}^{N_{\text{vis}}-2} \|\mathbf{KP}_{i+1}^{\text{3D}} - \mathbf{KP}_i^{\text{3D}}\|
\label{eq:m4}
\end{equation}
With 2--3 keypoints, a straight-line fallback is used. In the benchmark, full piecewise keypoint measurement ($N_{\text{vis}} \ge 4$) was achieved on 38 of 49 captures (78\%); the remaining captures used the straight-line fallback ($N_{\text{vis}} = 2$--3) or fell back to M3 when fewer than two endpoints passed the confidence and depth filters, and M4 returned no measurement on two captures (success rate 95.9\%). Table~\ref{tab:accuracy} reports the aggregate M4 results across all captures including these fallback cases. The reported MAPE of 5.51\% therefore reflects the full adaptive M4 pipeline rather than piecewise keypoint measurement alone.

Despite these improvements, M4's piecewise approximation still underestimates arc length: with $n$ chords spanning a total turn angle $\theta$ the residual bias is approximately $\theta^2/(24n^2)$, i.e.\ about 0.29\% for a $60^{\circ}$ arc sampled by four segments. It also depends on five regressed keypoints trained from $\sim 1500$ images and provides only sparse sampling. M5 addresses all three through dense medial-axis sampling, cubic spline fitting, and analytical arc-length integration.

\subsubsection{M5: Medial arc spline (proposed method)}
\begin{figure}[!t]
\centering
\includegraphics[width=0.75\linewidth]{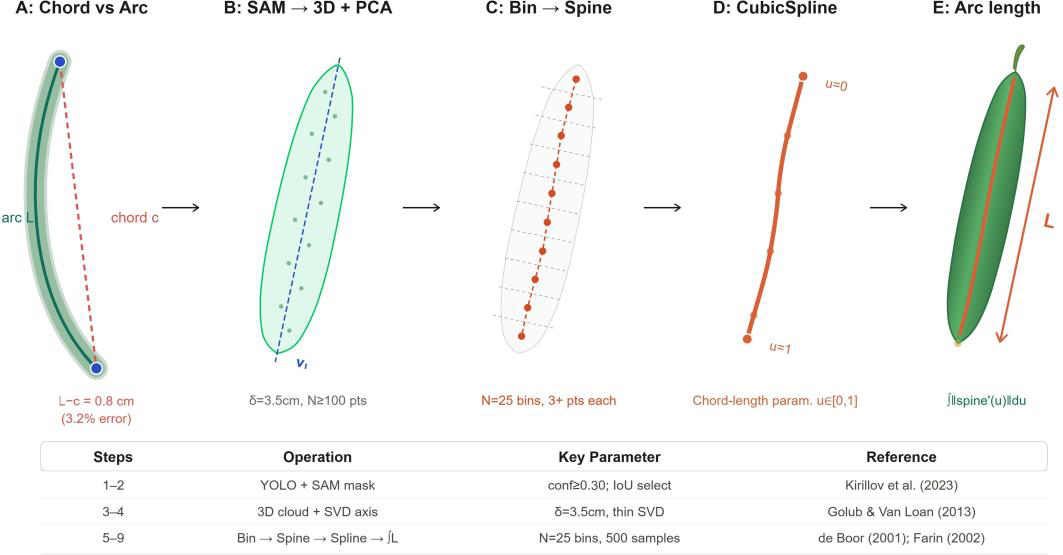}
\caption{M5 nine-step flowchart: YOLO26 detection, SAM segmentation, 3D cloud, SVD axis, cross-section binning, centroid spine, cubic spline fitting, arc-length integration, range validation.}
\label{fig:m5_spline}
\end{figure}

M1--M4 measure chords or piecewise approximations, systematically underestimating curved specimens. For a circular arc of length $L$ and radius $R$ the chord is $c = 2R \sin(L/2R)$, so the relative bias is $(L - c)/L \approx L^2/(24R^2)$; at the observed median $R \approx 42$~cm this is 1.5\% for a 25~cm fruit and 3.4\% for a 38~cm fruit. M5 computes physical arc length via a nine-step pipeline (Fig.~\ref{fig:m5_spline}). SAM-masked pixels are deprojected:
\begin{equation}
X = \frac{(u - c_x)\,z}{f_x}, \quad Y = \frac{(v - c_y)\,z}{f_y}, \quad Z = z
\label{eq:deproject}
\end{equation}
After median-depth filtering ($\delta = 3.5$~cm, at least 30 valid points), SVD yields $\mathbf{v}_1$. Points are binned into $N_b = 25$ cross-sections along $\mathbf{v}_1$; bin centroids form the spine, with the five extreme points at each end averaged to anchor the tips. Chord-length parameterisation maps vertebrae to $u \in [0, 1]$:
\begin{equation}
d_k = \|\mathbf{S}_{k+1} - \mathbf{S}_k\|, \quad u_k = \frac{\sum_{j<k} d_j}{d_{\text{total}}}
\label{eq:chord_param}
\end{equation}
Three cubic splines ($S''(0) = S''(1) = 0$) \citep{deboor2001, farin2002} are fitted for $x, y, z$. Arc length:
\begin{equation}
\hat{L}_{\text{M5}} = \int_0^1 \|\mathbf{S}'(u)\| \, du \approx \text{trapz}\bigl(\|[S'_x, S'_y, S'_z]\|,\; u\bigr)
\label{eq:arc_length}
\end{equation}
Range validation accepts $5 \le L \le 80$~cm. Under partial occlusion, affected bins are skipped and the spline interpolates across gaps.

\subsubsection{Adaptive method selector}
\begin{figure}[htbp]
\centering
\includegraphics[width=0.75\linewidth]{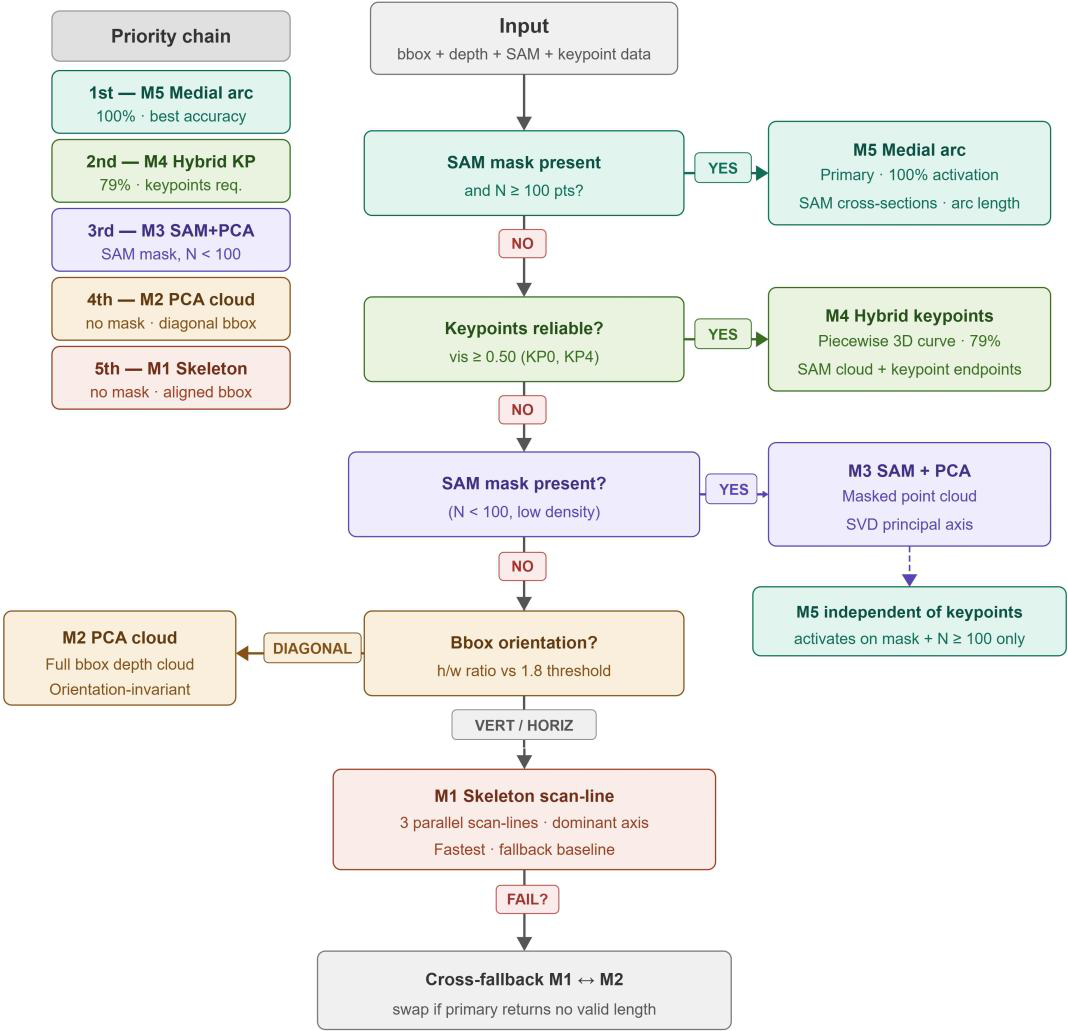}
\caption{Adaptive method-selection flowchart with cascading fallbacks from M5 (best) to M1 (fastest).}
\label{fig:selector_flowchart}
\end{figure}

Different cucumbers require different methods to calculate the length. Algorithm~\ref{alg:method_selection} formalises the adaptive method-selection pipeline illustrated in Fig.~\ref{fig:selector_flowchart}. For every detected cucumber instance the pipeline receives a bounding box, an aligned depth map, an optional SAM-refined mask, and five pose keypoints with associated visibility scores. The first decision gate checks whether a SAM mask is present and contains at least 30 valid depth points: if so, M5 (medial-arc spline) is selected as the primary method because it exploits the densest geometric information available. When, in addition, both endpoint keypoints (KP0 and KP4) exceed a visibility threshold of 0.50, M4 (hybrid keypoint) is computed in parallel for comparison, though M5 remains the reported measurement. If a SAM mask exists but its valid point count falls below 30, the pipeline falls back to M3 (SAM + PCA), which requires fewer inliers. When no SAM mask is available at all, the pipeline evaluates the bounding-box elongation ratio $r = \max(h,w)/\min(h,w)$: cucumbers whose bounding box is near-square ($r < 1.8$, typically diagonal specimens) are routed to M2 (PCA depth cloud), whereas elongated boxes ($r \ge 1.8$, vertical or horizontal orientations) are handled by M1 (skeleton scan-line). As a final safety net, a cross-fallback mechanism swaps M1 for M2 (or vice versa) whenever the initially selected method returns no valid length, ensuring that every detection produces a measurement whenever geometrically feasible.

\begin{algorithm}[!ht]
\caption{Adaptive method-selection pipeline.}
\label{alg:method_selection}
\small
\begin{algorithmic}[1]
\Require Bounding box $b_i$, depth map $D$, SAM mask $M_i$ (if available), keypoints KP0--KP4 with visibility $v_i$
\Ensure Length estimate $\hat{L}$ and selected method ID

\If{SAM mask $M_i$ available \textbf{and} $N_{\mathrm{valid}} \geq 30$}
    \State Apply \textbf{M5} (medial-arc spline) as \emph{primary}
    \If{$v_{\mathrm{KP0}} \geq 0.50$ \textbf{and} $v_{\mathrm{KP4}} \geq 0.50$}
        \State Compute \textbf{M4} (hybrid keypoint) for comparison
    \EndIf

\ElsIf{SAM mask $M_i$ available \textbf{and} $N_{\mathrm{valid}} < 30$}
    \State Apply \textbf{M3} (SAM + PCA)

\Else
    \State $r \gets \max(h,w) / \min(h,w)$
    \If{$r < 1.8$}
        \State Apply \textbf{M2} (PCA depth cloud)
    \Else
        \State Apply \textbf{M1} (skeleton scan-line)
    \EndIf
\EndIf

\Statex
\If{\textbf{M1} returns no valid length}
    \State Fall back to \textbf{M2}
\ElsIf{\textbf{M2} returns no valid length}
    \State Fall back to \textbf{M1}
\EndIf

\State \Return $\hat{L}$, method ID
\end{algorithmic}
\end{algorithm}

The full priority chain is $\text{M5} \rightarrow \text{M4} \rightarrow \text{M3} \rightarrow \text{M2} \rightarrow \text{M1}$, with a cross-fallback ($\text{M1} \leftrightarrow \text{M2}$) guaranteeing 100\% measurement coverage. The method ID, activation reason, and fallback path are logged per cucumber in the output CSV. On the capture benchmark, M5 was activated on 100\% of captures (all had SAM masks with $\ge 30$ valid depth points); M4 additionally processed 78\% of captures (38/49) where keypoint confidence was sufficient. Because the primary branch was selected for every capture, the M3, M2 and M1 fallback paths were not exercised on this benchmark; demonstrating them under controlled mask failure is identified as future work.

\section{Experimentation}
\label{sec:experimentation}
This section describes the experimental protocol for evaluating the non-contact cucumber length estimation framework, covering dataset, ground truth, metrics, and dashboard workflow.

\subsection{Dataset and annotations}
\begin{figure}[ht!]
\centering
\includegraphics[width=0.75\linewidth]{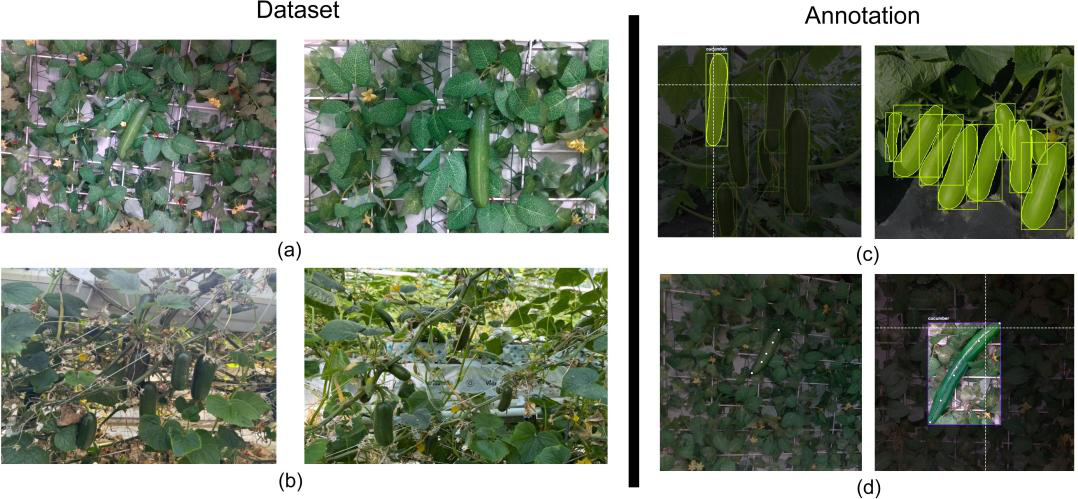}
\caption{Dataset samples: (a) laboratory, (b) greenhouse, (c) segmentation annotations, (d) keypoint annotations.}
\label{fig:dataset_samples}
\end{figure}

Images were collected from two environments (Fig.~\ref{fig:dataset_samples}): University of Windsor laboratory (uniform lighting) and commercial greenhouse (Prism farm) (variable illumination, dense canopy). Each image was annotated with instance segmentation masks and anatomical keypoints in Roboflow annotation software \citep{dwyer2026}.

Table~\ref{tab:dataset_structure} summarises the dataset: 1500 images, 4360 instances, 70:15:15 split, single cucumber class ($\sim 2.9$ instances per image).

\begin{table}[htbp]
\centering
\caption{Greenhouse cucumber dataset structure.}
\label{tab:dataset_structure}
\begin{tabular}{llll}
\toprule
\textbf{Split} & \textbf{Images} & \textbf{Instances} & \textbf{Ratio} \\ \midrule
Training & 1050 & 3045 & 70\% \\
Validation & 225 & 654 & 15\% \\
Test & 225 & 661 & 15\% \\ \midrule
Total & 1500 & 4360 & 100\% \\ \bottomrule
\end{tabular}
\end{table}

\subsection{Ground-truth collection protocol}
\begin{figure}[htbp]
\centering
\includegraphics[width=0.75\linewidth]{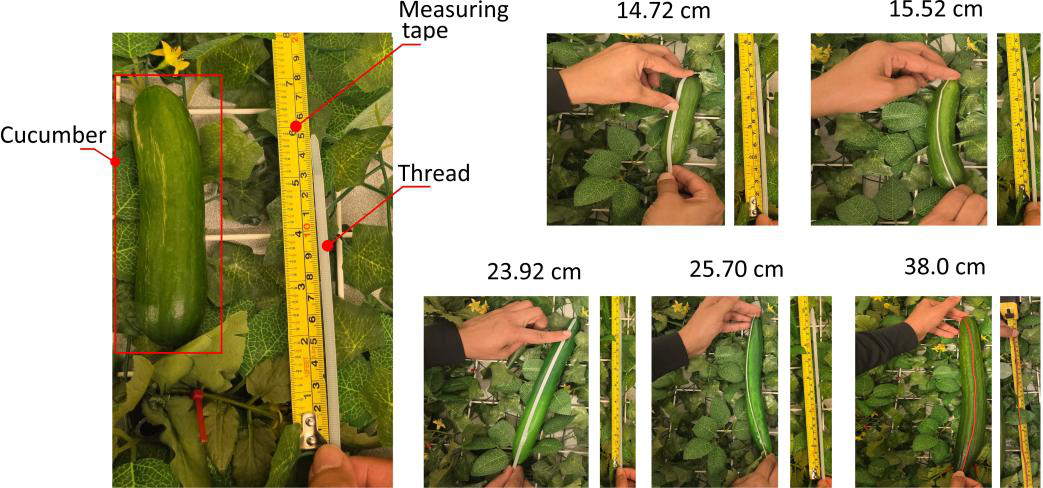}
\caption{Thread-based ground truth measurement protocol. Left: thread laid along dorsal midline and transferred to calibrated steel rule. Right: examples from the validation set spanning 14.72--38.0~cm, i.e.\ the medium to extra-large categories; the small category ($\sim 8$~cm) is represented only in the primary benchmark.}
\label{fig:ground_truth_protocol}
\end{figure}

Ground-truth lengths were measured using a flexible, non-stretchable thread laid along the dorsal (outer) midline of each cucumber from stem end to blossom end, then transferred to a calibrated steel rule (Fig.~\ref{fig:ground_truth_protocol}). This thread-tracing method captures the true arc length of the curved fruit surface, unlike a rigid vernier caliper which measures only the straight-line chord and therefore systematically underestimates curved specimens by the same chord-to-arc bias that motivates M5. The thread method has an operator variability of $\pm 1$--2~mm depending on how tightly the thread conforms to surface curvature.

\subsection{Evaluation metrics}
Each method is assessed against thread-based ground truth using four regression metrics. Let $\hat{L}_i$ denote predicted length, $L_{\text{GT},i}$ the ground truth, and $n$ the number of successful measurements:
\begin{equation}
\text{MAE} = \frac{1}{n} \sum_{i=1}^n |\hat{L}_i - L_{\text{GT},i}|
\end{equation}
\begin{equation}
\text{MAPE} = \frac{100}{n} \sum_{i=1}^n \frac{|\hat{L}_i - L_{\text{GT},i}|}{L_{\text{GT},i}}
\end{equation}
\begin{equation}
\text{RMSE} = \sqrt{\frac{1}{n} \sum_{i=1}^n (\hat{L}_i - L_{\text{GT},i})^2}
\end{equation}
\begin{equation}
R^2 = 1 - \frac{\sum_i (\hat{L}_i - L_{\text{GT},i})^2}{\sum_i (L_{\text{GT},i} - \bar{L}_{\text{GT}})^2}
\end{equation}

\subsection{Streamlit dashboard and operator workflow}
A proposed Streamlit dashboard integrates the capture pipeline, adaptive selector, and measurement output. The operator specifies row and position; the system executes burst capture, processes frames through YOLO26n + SAM, applies adaptive method selection, and performs 3D deduplication across stops. Annotated images, per-fruit lengths, and aggregate statistics are displayed; all data are exported as CSV and Excel as output of the Streamlit dashboard.

\section{Results and Deployment}
\label{sec:results}
This section presents quantitative and qualitative evaluation of M1--M5 on the benchmark dataset. All development used the capture benchmark exclusively; the five-specimen set of Section~\ref{subsec:heldout} was collected afterwards, with no parameter altered, and is reported as a held-out test set. Because the ground-truth distribution comprises three discrete size clusters, $R^2$ is dominated by between-cluster contrast and is negative within every cluster; MAE, RMSE and Bland--Altman limits of agreement are therefore the informative summaries and are used throughout.

\subsection{Detection and segmentation performance}
\begin{figure}[htbp]
\centering
\includegraphics[width=\linewidth]{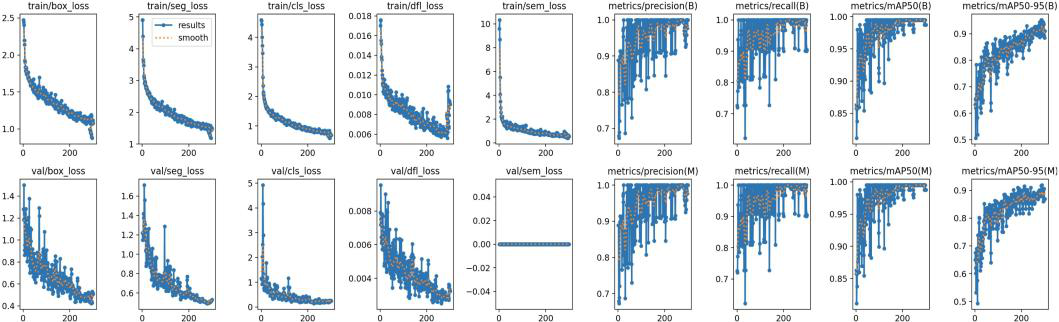}
\caption{YOLO26n training/validation over 300 epochs. Top: training losses and detection metrics. Bottom: validation losses and mask metrics.}
\label{fig:yolo_training}
\end{figure}

Fig.~\ref{fig:yolo_training} shows YOLO26n training over 300 epochs. All losses converge monotonically without train--validation divergence. Detection: precision and recall $>0.90$ by epoch 100; mAP@50(B) $\approx 0.95$, mAP@50-95(B) $\approx 0.90$. Segmentation: mAP@50(M) $\approx 0.96$, mAP@50-95(M) $>0.85$. Near parity between box and mask metrics is consistent with high-fidelity masks suitable for SAM refinement.

\subsection{Length measurement accuracy}
\begin{figure}[htbp]
\centering
\includegraphics[width=\linewidth]{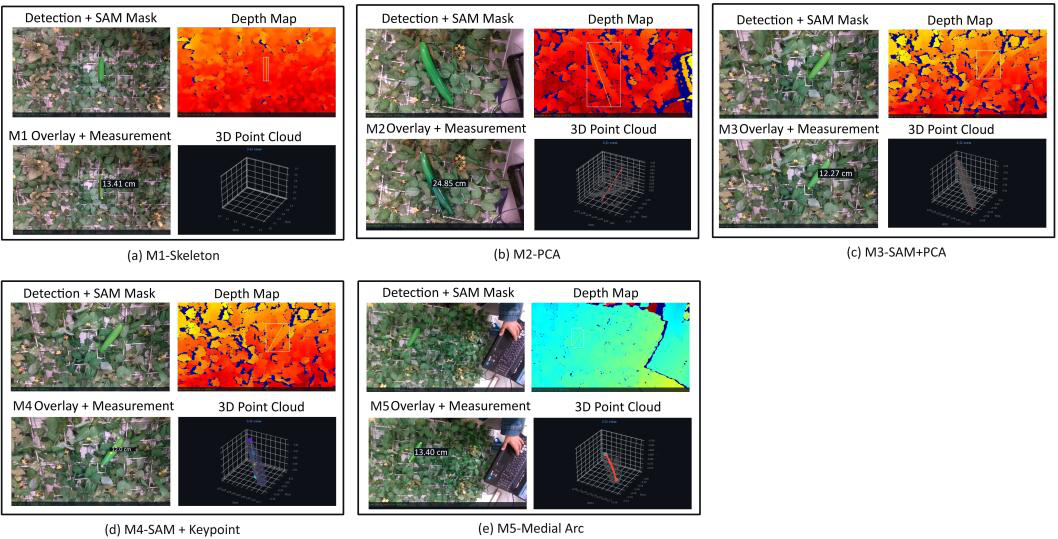}
\caption{Qualitative comparison: detection + SAM mask, depth map, measurement overlay, and 3D point cloud for each method. (a) M1: 13.41~cm, (b) M2: 24.85~cm, (c) M3: 12.27~cm, (d) M4: 17.0~cm, (e) M5: 13.40~cm.}
\label{fig:qualitative_comparison}
\end{figure}

Fig.~\ref{fig:qualitative_comparison} compares all five methods qualitatively: M1 (13.41~cm, vertical specimen), M2 (24.85~cm, SVD axis on oblique fruit), M3 (12.27~cm, mask-constrained), M4 (17.0~cm, piecewise keypoints), and M5 (13.40~cm, spline tracing curvature). Progressive geometric improvement is visible in the 3D views.

Table~\ref{tab:accuracy} reports quantitative metrics. M5 achieves the lowest error: MAE 0.58~cm, RMSE 0.89~cm, MAPE 4.13\% (95\% CI: 2.15--6.00\%). M1 shows the highest error (MAPE 9.68\%). Because the 49 captures derive from only seven physical specimens, observations are clustered and capture-level tests overstate the effective sample size; inference therefore uses a paired specimen-clustered bootstrap (10,000 resamples of specimens). On this basis M5 achieves significantly lower MAE than M1 ($\Delta$MAE 1.04~cm, 95\% CI [0.28, 1.70]) and M2 (0.25~cm, [0.02, 0.46]), while its advantage over M3 (0.29~cm, [$-$0.03, 0.56]) and M4 (0.28~cm, [$-$0.03, 0.55]) is consistent in sign but not statistically conclusive at seven specimens. We note that a specimen-level Wilcoxon signed-rank test cannot attain the Bonferroni-corrected threshold $\alpha = 0.0125$ at $n = 7$, since its minimum attainable two-sided $p$-value is 0.0156; the bootstrap is used in preference for this reason. Per-size-category results are given in Table~\ref{tab:accuracy_by_size} and bias and precision in Table~\ref{tab:bias_precision}.

\begin{table}[htbp]
\centering
\caption{Accuracy comparison. Best in bold. Specimen-clustered bootstrap 95\% CIs on MAE: M1 [0.85, 2.32], M2 [0.54, 1.07], M3 [0.53, 1.15], M4 [0.51, 1.16], M5 [0.44, 0.69]~cm. $R^2$ is retained for continuity with prior work but is inflated by the three-cluster ground-truth distribution and is negative within every cluster (Section~\ref{subsec:limitations}); it should not be used to rank methods.}
\label{tab:accuracy}
\footnotesize
\setlength{\tabcolsep}{3pt}
\begin{tabularx}{\linewidth}{@{}l >{\raggedright\arraybackslash}X c c c c@{}}
\toprule
\textbf{Method} & \textbf{Description} & \makecell{\textbf{MAE}\\\textbf{(cm)}} & \makecell{\textbf{RMSE}\\\textbf{(cm)}} & \makecell{\textbf{MAPE}\\\textbf{(\%)}} & \textbf{$R^2$} \\ \midrule
M1 & Skeleton scan-line & 1.63 & 2.26 & 9.68 & 0.89 \\
M2 & PCA depth cloud & 0.83 & 1.12 & 5.31 & \textbf{0.98} \\
M3 & SAM + PCA & 0.86 & 1.02 & 5.82 & 0.97 \\
M4 & Hybrid keypoint & 0.85 & 1.01 & 5.51 & 0.97 \\
\textbf{M5} & \textbf{Medial arc spline} & \textbf{0.58} & \textbf{0.89} & \textbf{4.13} & \textbf{0.98} \\ \bottomrule
\end{tabularx}
\end{table}

\begin{table}[htbp]
\centering
\caption{Accuracy by size category. MAE\,/\,RMSE in cm, MAPE in \%. The $\ge 20$~cm band is the commercially graded range and derives from two specimens (24.5 and 25.0~cm).}
\label{tab:accuracy_by_size}
\scriptsize
\setlength{\tabcolsep}{3pt}
\begin{tabularx}{\linewidth}{@{}l *{3}{>{\centering\arraybackslash}X}@{}}
\toprule
\textbf{Method} & \makecell{\textbf{Small}\\\textbf{$<$12\,cm}\\($n{=}20$)} & \makecell{\textbf{Medium}\\\textbf{12--20\,cm}\\($n{=}20$)} & \makecell{\textbf{Large}\\\textbf{$\ge$20\,cm}\\($n{=}20$)} \\
\midrule
M1 & 0.76/0.88/10.50 & 0.98/1.19/7.49 & 2.85/3.41/11.49 \\
M2 & 0.38/0.47/5.27  & 0.73/0.89/5.65 & 1.23/1.55/4.97  \\
M3 & 0.72/0.80/10.15 & 0.49/0.67/3.74 & 1.33/1.38/5.36  \\
M4 & 0.63/0.64/8.69  & 0.52/0.71/3.96 & 1.33/1.39/5.38  \\
M5 & 0.35/0.46/4.94  & 0.66/1.13/5.09 & 0.64/0.80/2.59  \\
\bottomrule
\end{tabularx}
\end{table}

Table~\ref{tab:accuracy_by_size} breaks down measurement accuracy by cucumber size category. M5 achieves the lowest MAE and MAPE in both the small ($<$12~cm) and large ($\ge$20~cm) categories, with MAE/RMSE/MAPE of 0.35/0.46/4.94 and 0.64/0.80/2.59, respectively. Its advantage is most pronounced for large, commercially graded cucumbers, where it reduces MAE by 78\% relative to M1 (0.64 vs.\ 2.85~cm) and MAPE by 48\% relative to M2 (2.59 vs.\ 4.97\%). In the medium category (12--20~cm), M3 and M4 marginally outperform M5 (MAE 0.49 and 0.52~cm vs.\ 0.66~cm), suggesting that PCA- and keypoint-based methods can be competitive when cucumber curvature is moderate. M1 consistently shows the highest errors across all size bands, with MAPE exceeding 7\% in every category and reaching 11.49\% for large specimens, confirming that the skeleton scan-line approach struggles with longer, more curved fruit. M2 performs well on small cucumbers (MAE 0.38~cm) but its accuracy degrades for larger specimens. Overall, the size-stratified results reinforce that M5 generalises most reliably across the full range of cucumber lengths encountered in greenhouse production.

\begin{table}[htbp]
\centering
\caption{Bias and precision. Bland--Altman bias and 95\% limits of agreement (predicted $-$ ground truth), and within-specimen repeatability expressed as the pooled standard deviation of predicted length across repeated captures of the same fruit.}
\label{tab:bias_precision}
\footnotesize
\setlength{\tabcolsep}{3pt}
\begin{tabularx}{\linewidth}{@{}l c >{\centering\arraybackslash}X c c@{}}
\toprule
\textbf{Method} & \makecell{\textbf{Bias}\\\textbf{(cm)}} & \makecell{\textbf{95\% LoA}\\\textbf{(cm)}} & \makecell{\textbf{Within-spec.}\\\textbf{SD (cm)}} & \makecell{\textbf{Mean}\\\textbf{CV (\%)}} \\
\midrule
M1 & $-$1.56 & [$-$4.79, +1.68] & 1.24 & 8.45 \\
M2 & $-$0.21 & [$-$2.38, +1.96] & 1.02 & 6.46 \\
M3 & $-$0.85 & [$-$1.95, +0.24] & 0.41 & 3.34 \\
M4 & $-$0.85 & [$-$1.94, +0.24] & 0.75 & 5.22 \\
\textbf{M5} & \textbf{+0.08} & \textbf{[$-$1.66, +1.83]} & \textbf{0.37} & \textbf{2.51} \\
\bottomrule
\end{tabularx}
\end{table}

Table~\ref{tab:bias_precision} summarises bias and repeatability for the five measurement methods. M5 exhibits the smallest systematic error, with a Bland--Altman bias of only $+$0.08~cm, indicating near-zero average over- or under-estimation relative to the thread-based ground truth. Its 95\% limits of agreement [$-$1.66, $+$1.83]~cm are also the narrowest overall span (3.49~cm), confirming that individual predictions deviate least from ground truth. M5 further achieves the best within-specimen repeatability, with a pooled standard deviation of 0.37~cm and a mean coefficient of variation of 2.51\%, meaning that repeated captures of the same cucumber yield highly consistent length estimates. By contrast, M1 shows the largest bias ($-$1.56~cm), widest limits of agreement (6.47~cm span), and poorest repeatability (SD 1.24~cm, CV 8.45\%), reflecting its sensitivity to bounding-box alignment and scan-line placement. M2 achieves a relatively low bias ($-$0.21~cm) but exhibits moderate repeatability (SD 1.02~cm), while M3 and M4 share identical bias ($-$0.85~cm) and narrow limits of agreement but with intermediate repeatability. Overall, these results confirm that the medial-arc spline approach (M5) provides the most accurate and repeatable length measurements among the five methods evaluated.

\subsection{Comparative analysis}
\begin{figure}[htbp]
\centering
\includegraphics[width=\linewidth]{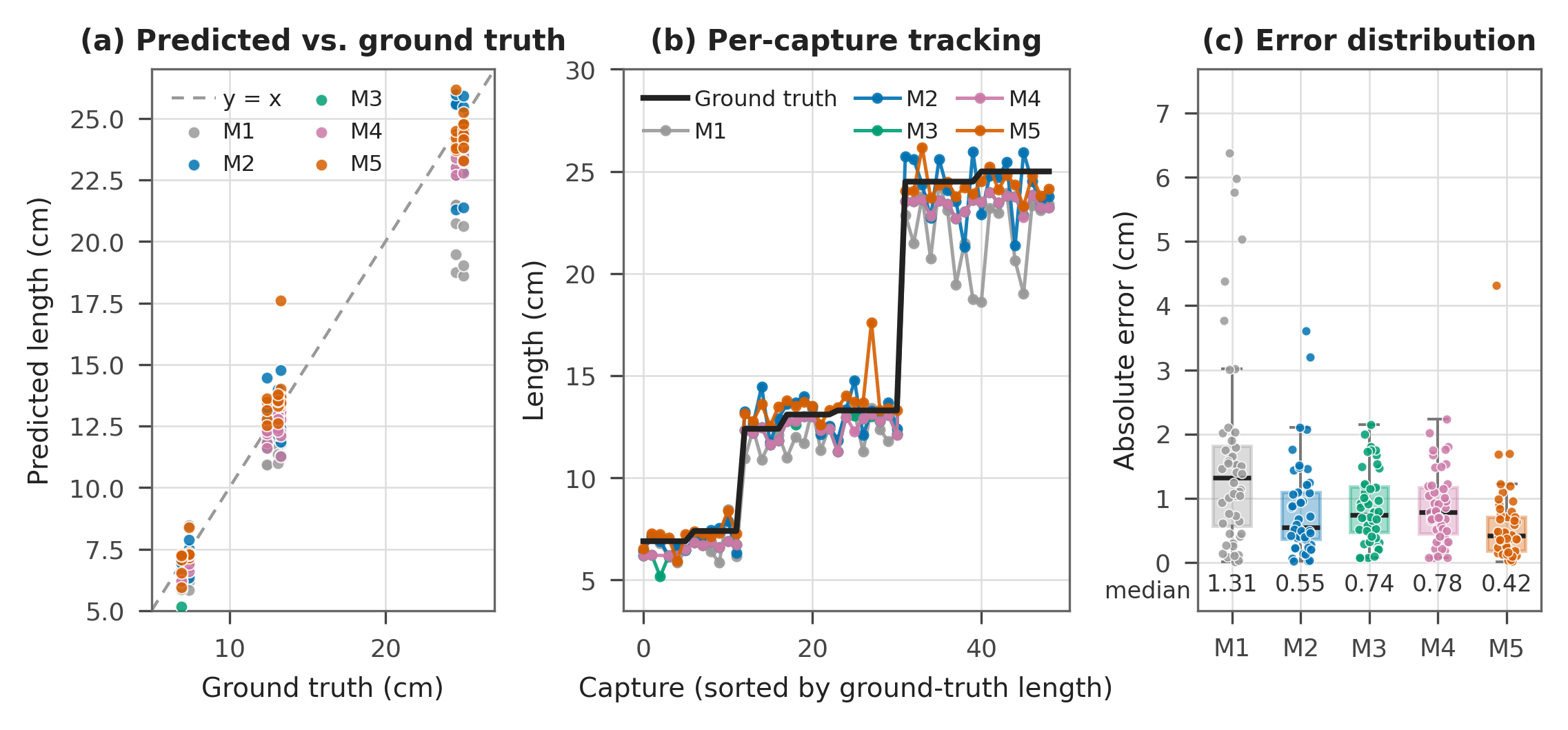}
\caption{Method comparison on the 49-capture benchmark. (a) Predicted vs.\ ground truth. (b) Per-capture tracking, captures sorted by ground-truth length. (c) Error distribution (box, individual captures overlaid). Median absolute error: M1 = 1.31, M2 = 0.55, M3 = 0.74, M4 = 0.78, M5 = 0.42~cm.}
\label{fig:comparative_analysis}
\end{figure}

Fig.~\ref{fig:comparative_analysis} presents three views. The scatter plot~(a) shows M5 clustering tightest along $y = x$ across all sizes; M1 underestimates large specimens. Per-capture tracking~(b) confirms M5 and M4 follow ground truth most faithfully; SAM-based methods outperform bbox-only for medium/large cucumbers. The error distribution~(c) shows monotonically decreasing median from M1 (1.31~cm) to M5 (0.42~cm) with narrowing IQR; M5 produces the fewest large errors, with 1 of 49 captures exceeding 2~cm against 11 for M1, although a single spline-overshoot failure of 4.31~cm remains and is analysed in Section~\ref{subsec:method_analysis}.

\begin{figure}[!t]
\centering
\includegraphics[width=\linewidth]{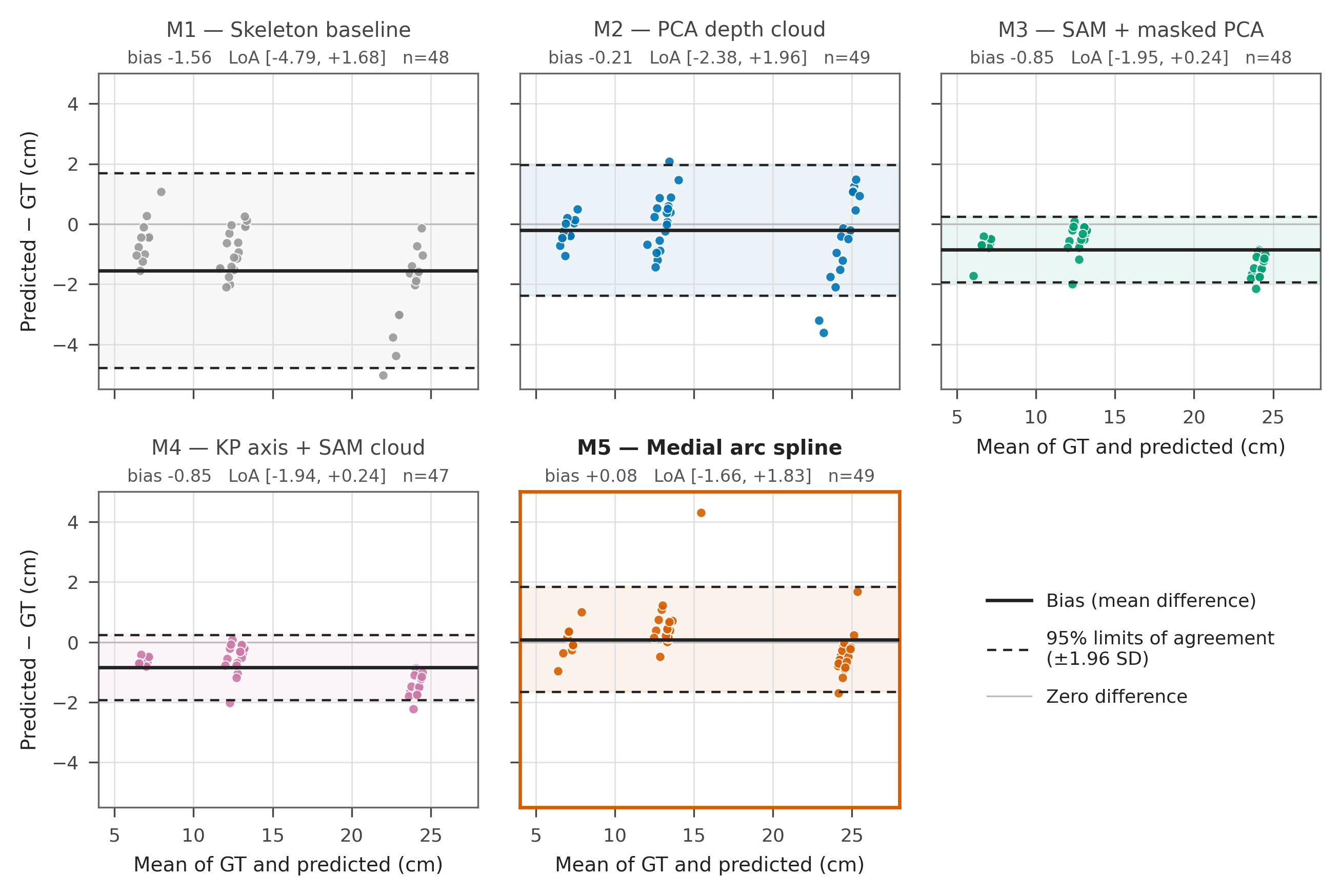}
\caption{Bland--Altman agreement against thread ground truth for all five methods. Solid line, bias (mean difference); dashed lines, 95\% limits of agreement ($\pm$1.96 SD); shaded band, the interval between them. The $y$-axis is shared across panels. M5 is the only method whose bias is not significantly different from zero ($+$0.08~cm, $p = 0.51$); M3 and M4 carry a systematic underestimate of 0.85~cm. The M5 point near $+$4.3~cm is the spline-overshoot failure of Section~\ref{subsec:method_analysis}.}
\label{fig:bland_altman}
\end{figure}

\begin{figure}[!t]
\centering
\includegraphics[width=\linewidth]{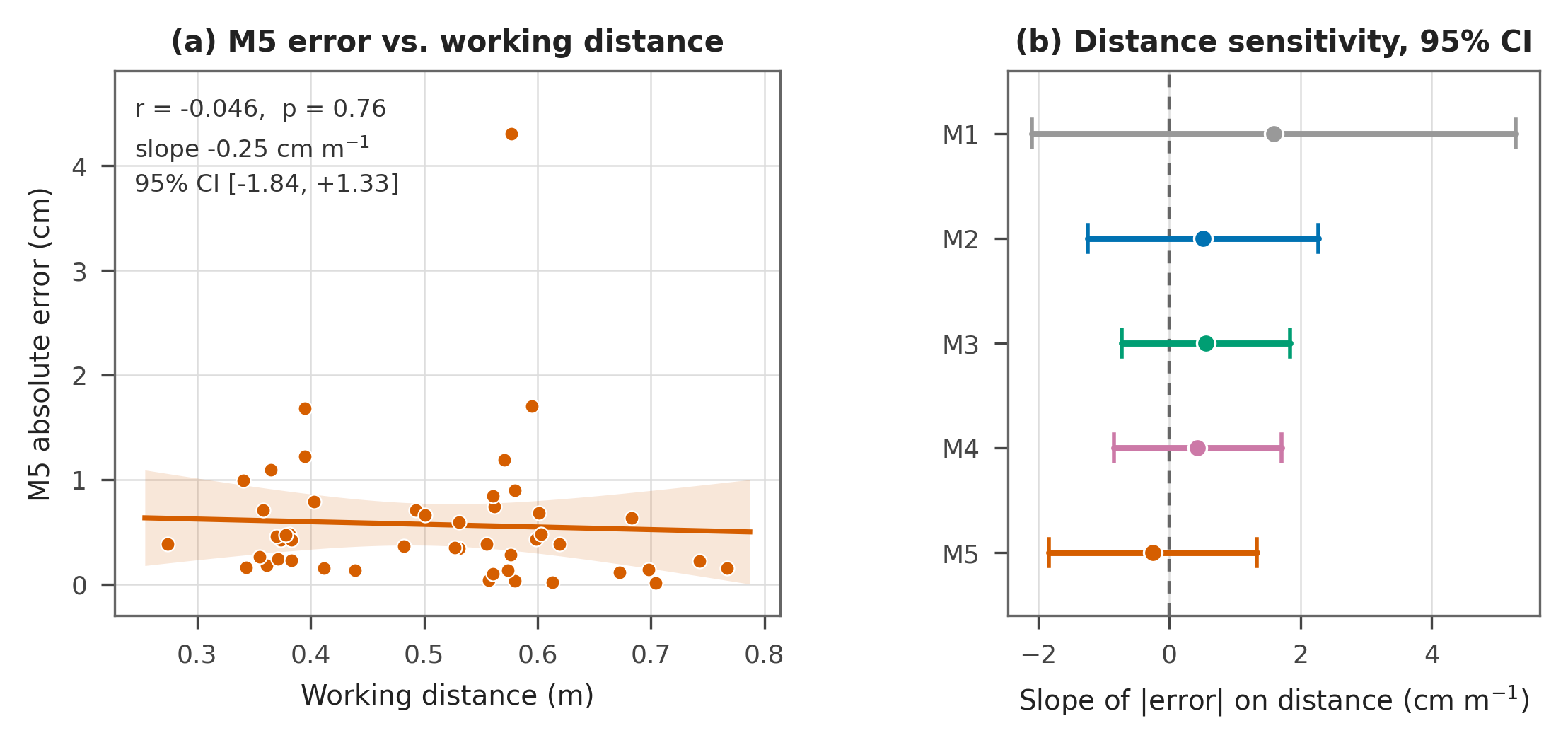}
\caption{Measurement error against camera working distance over the 0.27--0.77~m operating range. (a) M5 absolute error per capture with ordinary least-squares fit and 95\% confidence band; the slope is $-0.25$~cm\,m$^{-1}$ ($r = -0.046$, $p = 0.76$), i.e.\ no detectable degradation with distance. (b) Slope of absolute error on working distance for all five methods with 95\% confidence intervals; every interval includes zero. Working distance and specimen length are uncorrelated across the benchmark ($r = -0.029$, $p = 0.85$), so the comparison is not confounded by fruit size.}
\label{fig:working_distance}
\end{figure}

Fig.~\ref{fig:bland_altman} shows the Bland--Altman agreement for all five methods. M5 is the only estimator without significant bias in this benchmark: its Bland--Altman bias of $+$0.08~cm is not significantly different from zero (one-sample $t$-test, $p = 0.51$), whereas M3 and M4 carry a systematic underestimate of 0.85~cm. The precision ordering differs from the accuracy ordering: M3 and M4 are the most repeatable (within-specimen SD 0.41 and 0.37~cm) but biased, while M5 is unbiased at the cost of higher per-capture variance (0.75~cm). For census-scale length distributions, where many fruit are aggregated, low bias dominates; for repeated measurement of an individual fruit, the precision of M3 and M4 is preferable. Increasing the burst depth or temporal smoothing via SAM~2 is the natural route to reducing M5 variance. Error magnitude showed no significant dependence on working distance over the 0.27--0.77~m operating range for any method (Fig.~\ref{fig:working_distance}; M5: Pearson $r = -0.046$, $p = 0.76$, slope $-0.25$~cm\,m$^{-1}$, 95\% CI [$-$1.84, $+$1.33]). Working distance is uncorrelated with specimen length across the benchmark ($r = -0.029$, $p = 0.85$), so this is not a size confound.

\subsection{Held-out test set}
\label{subsec:heldout}
\begin{figure}[htbp]
\centering
\includegraphics[width=\linewidth]{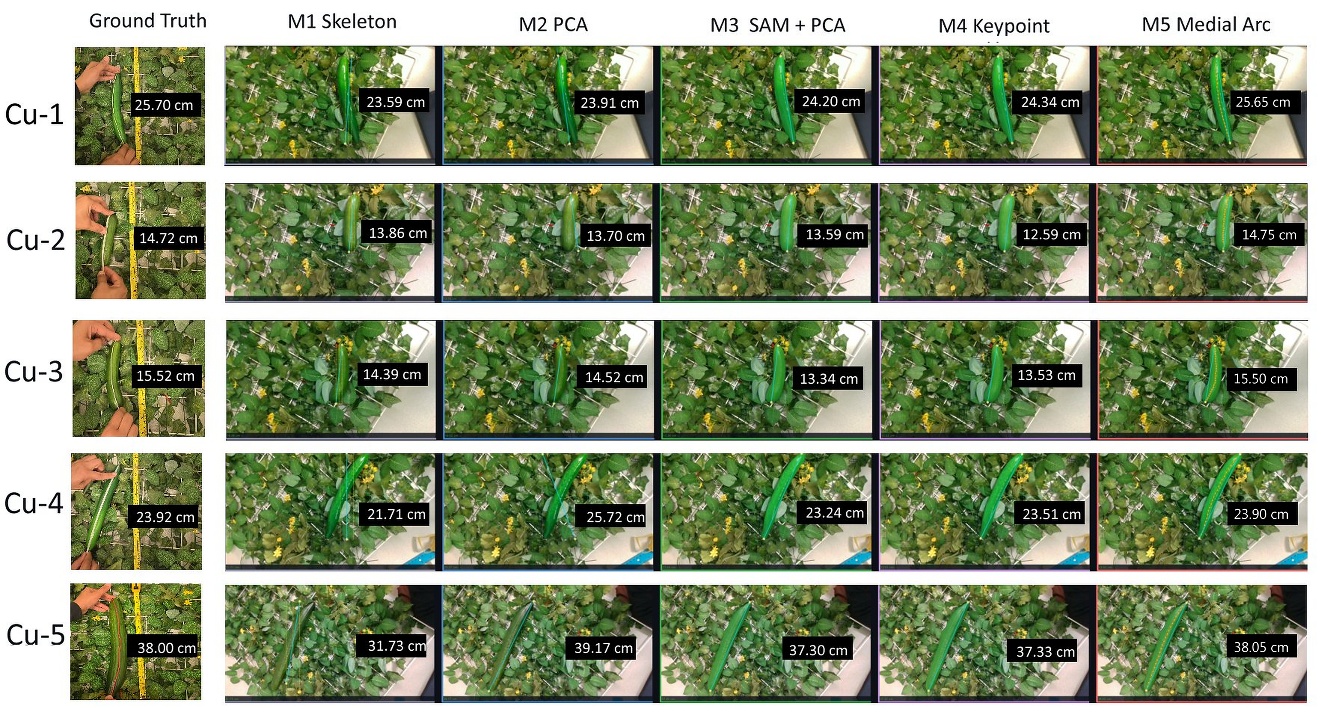}
\caption{Five-method comparison on the five-specimen held-out test set (GT 14.72--38.00~cm). Each row shows the same cucumber processed by M1 (skeleton), M2 (PCA), M3 (SAM+PCA), M4 (keypoints), and M5 (medial arc). M5 agrees with the thread-based ground truth to within 0.05~cm for all five specimens.}
\label{fig:heldout_comparison}
\end{figure}

To validate method behaviour on specimens larger than the primary benchmark ($\ge 25.0$~cm) and in the uncharacterised 13.3--24.5~cm range, a separate held-out test set of five individual cucumbers was collected after the primary benchmark and after all method parameters were fixed; no threshold was tuned on it. It was collected with ground-truth lengths of 14.72, 15.52, 23.92, 25.70, and 38.00~cm measured by the thread protocol (Fig.~\ref{fig:ground_truth_protocol}). All five methods were applied to each specimen under identical capture conditions. Table~\ref{tab:heldout} summarises the results. The validation results are illustrated in Fig.~\ref{fig:heldout_comparison}.

\subsection{Deployment: CucumberVision dashboard}
The pipeline is deployed as CucumberVision, a Streamlit app (Fig.~\ref{fig:dashboard}) packaging all detection, segmentation, and measurement stages. The capture interface~(a) provides configurable parameters, annotated frames, and per-fruit listings. The results tab~(b) summarises session statistics, provides a measurement table with CSV/Excel export, and confirms method distribution. Live RealSense streaming and offline image upload are both supported. The proposed dashboard will be deployed on the Edge device (Raspberry Pi or Jetson Nano) as future work. The dashboard covers a hand-positioned camera. Section~\ref{sec:simulation} moves the same measurement chain onto robot wrists, where the measured length decides where each harvested fruit is placed.

\section{Simulation-Based Robotic Harvesting and Grading}
\label{sec:simulation}
\begin{figure}[!t]
\centering
\includegraphics[width=\linewidth]{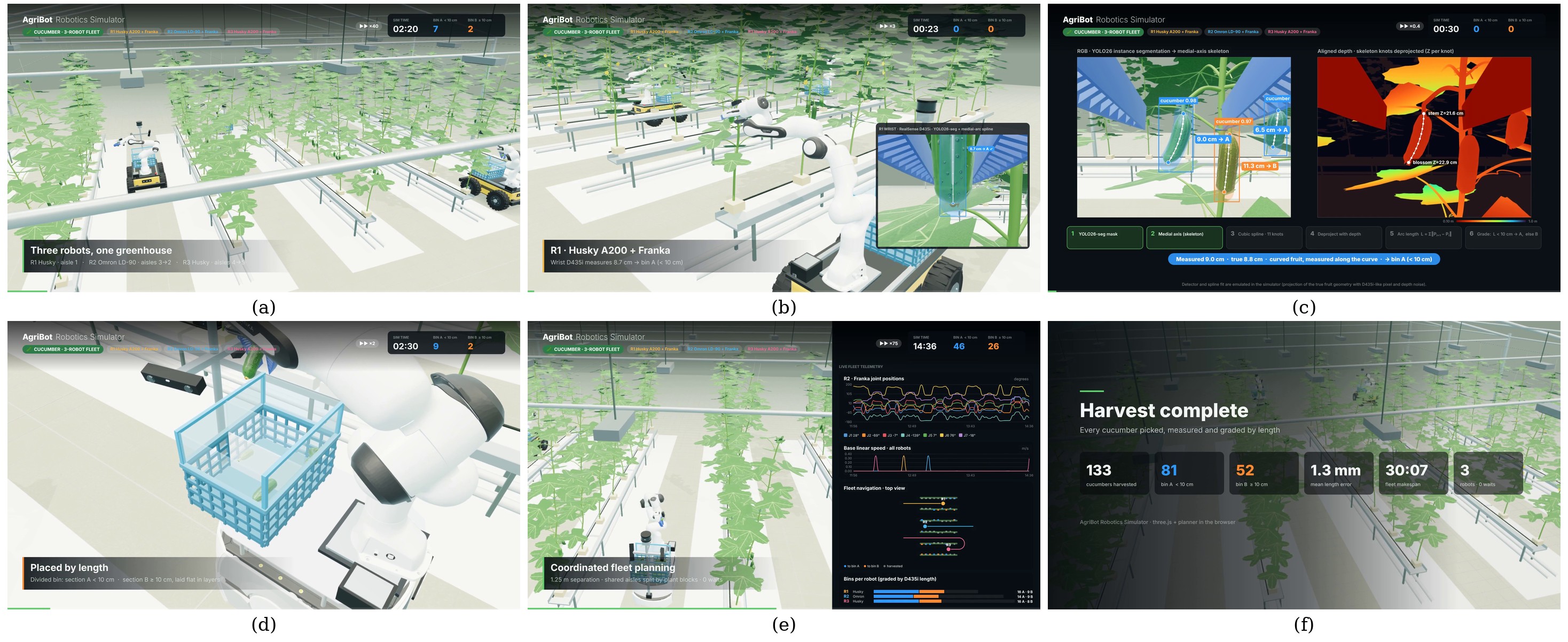}
\caption{Simulated multi-robot cucumber harvesting and length grading in the AgriBot Robotics Simulator. (a)~Three-robot fleet in a six-row high-wire greenhouse: R1 (Husky A200 + Franka) in aisle~1, R2 (Omron LD-90 + Franka) in aisles 3$\rightarrow$2, R3 (Husky A200 + Franka) in aisles 4$\rightarrow$5. (b)~Wrist-mounted RealSense D435i view at the pre-grasp pose, with the M5 length and bin assignment overlaid (8.7~cm $\rightarrow$ bin~A). (c)~In-simulation measurement chain: instance mask and medial-axis skeleton on RGB (left), skeleton knots deprojected on the aligned depth image (right), and the six-step pipeline ending in the grading rule. A curved fruit of true length 8.8~cm measures 9.0~cm along the curve. (d)~Flat placement into the divided on-board bin (section~A $<$10~cm, section~B $\ge$10~cm). (e)~Live fleet telemetry: Franka joint positions, base linear speed, top-view fleet navigation and per-robot bin counts. (f)~Run summary: 133 fruit harvested, 81 in bin~A and 52 in bin~B, 1.3~mm mean length error against simulated ground truth, and a 30:07 (min:s) simulated fleet makespan with no coordination waits.}
\label{fig:simulation}
\end{figure}

Sections~\ref{sec:methodology}--\ref{sec:results} evaluate M5 as a stand-alone instrument. In commercial use, length estimation would more likely run on a harvesting robot that measures each fruit just before picking and uses the result to decide where to place it. To show this integration and study it at fleet scale before field trials, we embedded the M5 measurement chain in the AgriBot Robotics Simulator. This is a browser-based greenhouse simulator in which rendering (three.js) and task and motion planning run in the browser (Fig.~\ref{fig:simulation}).

\subsection{Greenhouse scene and crop model}
The scene reproduces a six-row high-wire cucumber greenhouse with 54 plants on raised gutters, separated by aisles wide enough for a mobile manipulator. Each plant carries two to three fruit of 6--13~cm. About one third of the fruit are curved, so that the chord-to-arc difference that motivates M5 (Section~\ref{sec:methodology}) appears in the simulated population. Fruit are rendered with warty, glossy skin, and flowers and leaves partially occlude the fruit in the wrist-camera view. Because every fruit is generated from a known parametric centre-line, its true arc length is available exactly. This gives a noise-free reference, unlike the $\pm 1$--2~mm uncertainty of the thread protocol (Section~\ref{sec:experimentation}).

\subsection{Robot fleet and coordinated planning}
The fleet comprises three mobile manipulators. R1 and R3 are Clearpath Husky A200 skid-steer platforms and R2 is an Omron LD-90 autonomous mobile robot (AMR); each carries a Franka arm. Every arm has a RealSense D435i on the wrist, a soft Fin Ray gripper, and a divided, high-sided collection bin on the base. Initial aisles and entry ends are assigned per robot (Fig.~\ref{fig:simulation}a). The planner generates serpentine routes through the rows. It splits aisles shared between robots into plant blocks and plans reverse-out exits at row ends. It also enforces a minimum inter-robot separation of 1.25~m, with priority-based waiting when two routes would violate it. Joint trajectories, base speed and fleet positions are logged continuously and can be exported as CSV and PNG (Fig.~\ref{fig:simulation}e).

\subsection{In-the-loop length measurement}
At each fruit, the arm moves to a pre-grasp pose that frames the fruit in the wrist camera, and the length is fixed at this view before the grasp begins. The in-simulation chain follows M5 (Section~\ref{sec:methodology}) in six steps (Fig.~\ref{fig:simulation}c):
\begin{enumerate}[nosep,leftmargin=*]
    \item an instance mask is obtained for the fruit;
    \item the medial axis (skeleton) of the mask is extracted from stem to blossom end;
    \item a cubic spline with 11 knots is fitted along the skeleton;
    \item the knots are deprojected to 3D with the aligned depth image using Eq.~(\ref{eq:deproject});
    \item arc length is accumulated as $\hat{L} = \sum_i \|\mathbf{P}_{i+1} - \mathbf{P}_i\|$ over the densely sampled spline;
    \item the fruit is graded by $\hat{L} < 10$~cm $\rightarrow$ bin~A, otherwise bin~B.
\end{enumerate}
Two simplifications relative to the physical pipeline should be stated. First, the detector and spline fit are \emph{emulated}: the mask and depth are generated by projecting the true fruit geometry through a D435i-like camera model with pixel and depth noise, rather than by running the trained YOLO26n and SAM models on rendered images. Second, the spine is represented by 11 spline knots rather than the $N_b = 25$ cross-section bins used on real data. The 10~cm threshold suits the simulated 6--13~cm population; commercial class boundaries such as 20 or 25~cm (Section~\ref{sec:introduction}) are set the same way.

\subsection{Harvest-and-grade cycle}
After measurement, the Fin Ray gripper grasps the fruit and the peduncle is cut. The fruit is then laid flat in the bin section given by its grade and stacked in layers within each section (Fig.~\ref{fig:simulation}d). The base advances to the next plant block once every reachable fruit on the current plant has been processed. Length grading therefore adds no extra camera viewpoint or base stop to the harvest cycle, because the measurement uses the pre-grasp view the grasp planner already requires.

\subsection{Simulation results}
Table~\ref{tab:simulation} summarises a complete fleet run. All 133 fruit in the scene were harvested and graded: 81 into bin~A ($<$10~cm) and 52 into bin~B ($\ge$10~cm). Against the exact simulated ground truth, the mean absolute length error was 1.3~mm. The fleet finished in 30:07 (min:s) of simulated time, about 4.4 fruit per minute for the fleet or roughly 1.5 per robot, and the separation constraint never forced a coordination wait. The per-robot workload was balanced: part-way through the run R1, R2 and R3 had graded 25, 23 and 24 fruit respectively (Fig.~\ref{fig:simulation}e).

\begin{table}[htbp]
\centering
\caption{Simulated three-robot harvest-and-grade run. Length error is measured against the exact simulated centre-line length.}
\label{tab:simulation}
\footnotesize
\begin{tabularx}{\linewidth}{@{}X r@{}}
\toprule
\textbf{Quantity} & \textbf{Value} \\ \midrule
Robots (Husky A200 / Omron LD-90, each + Franka) & 2 / 1 \\
Greenhouse rows / plants & 6 / 54 \\
Fruit per plant; fruit length & 2--3; 6--13~cm \\
Inter-robot separation constraint & 1.25~m \\
Fruit harvested and graded & 133 \\
Bin~A ($<$10~cm) / Bin~B ($\ge$10~cm) & 81 / 52 \\
Mean absolute length error & 1.3~mm \\
Fleet makespan (simulated) & 30:07 min:s \\
Fleet throughput (derived) & $\approx$4.4 fruit\,min$^{-1}$ \\
Coordination waits & 0 \\ \bottomrule
\end{tabularx}
\end{table}

\subsection{Interpretation and scope}
The 1.3~mm simulated error is not comparable with the 5.8~mm MAE measured on real fruit (Table~\ref{tab:accuracy}). It shows only that the spline arc-length chain is consistent with the true geometry when the D435i-like pixel and depth noise model is the only error source. The error sources that dominate on real data are absent by construction: SAM mask bleed (Fig.~\ref{fig:m5_failure}), detector misses under occlusion, and ground-truth uncertainty. The simulation therefore supports three narrower claims. First, M5 can serve as the grading signal in a pick-and-place cycle without adding sensing poses. Second, its output can drive bin assignment across a heterogeneous fleet of skid-steer and AMR bases. Third, the fleet planner can coordinate several robots in shared aisles without waits at this density. Closing the gap to real accuracy requires running the trained detector and SAM on rendered images inside the loop, which is left to future work.

\section{Discussion}
\label{sec:discussion}

\subsection{Summary of key findings}
M5 achieves the lowest error on all metrics (MAE 0.58~cm, MAPE 4.13\%) and is the only method without significant bias in this benchmark (Bland--Altman bias $+$0.08~cm); the accuracy ranking replicates in a second capture condition (Section~\ref{subsec:heldout}). Under specimen-clustered bootstrap inference it significantly outperforms M1 and M2, with a consistent but not conclusive advantage over M3 and M4. Ranked by MAPE the ordering is M1 (9.68\%), M3 (5.82\%), M4 (5.51\%), M2 (5.31\%), M5 (4.13\%). Accuracy therefore improves overall from the simplest to the most geometrically explicit method, but not monotonically in method complexity: M3 falls behind M2 because SAM mask bleed on small specimens displaces the principal axis (Section~\ref{subsec:method_analysis}). The adaptive selector ensures 100\% coverage. In simulation, the same arc-length chain served as the per-fruit grading signal for a three-robot harvesting fleet (Section~\ref{sec:simulation}), grading all 133 fruit with no extra sensing poses. Deprojection uses colour-stream intrinsics throughout, which is required after alignment to the colour stream; the alternative inflates measured length by approximately one third (Section~\ref{sec:methodology}).

\subsection{Method-specific analysis}
\label{subsec:method_analysis}
\begin{figure}[!t]
\centering
\includegraphics[width=0.9\linewidth]{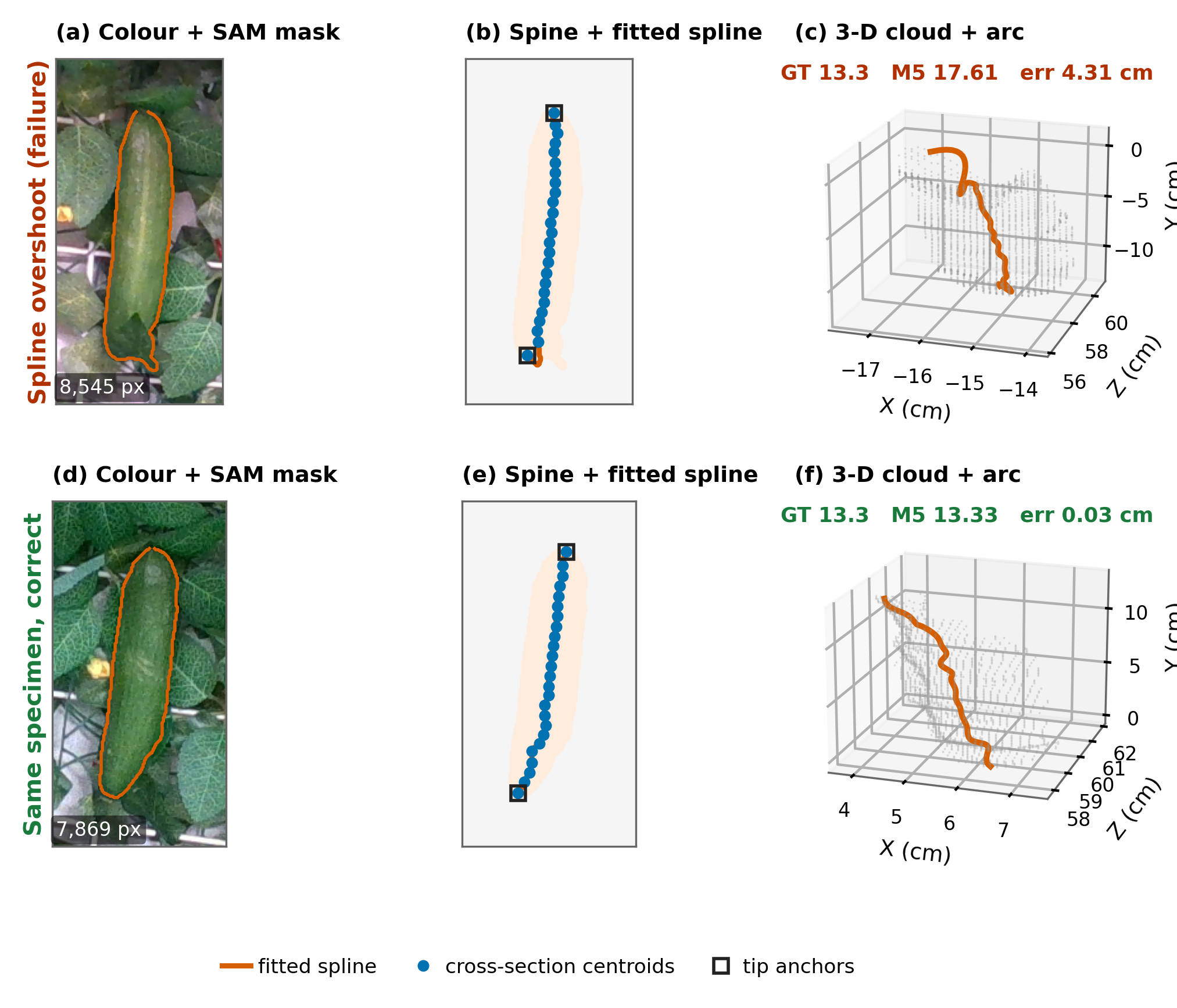}
\caption{The dominant M5 failure mode, shown against a control. Both rows are the same specimen (ground truth 13.3~cm) captured seconds apart at the same working distance (0.58~m) and orientation; only the SAM mask differs. Top row: the mask bleeds past the blossom end into the leaf litter behind the fruit~(a), the cross-section centroids follow the bleed and the fitted spline hooks outwards with them~(b,~c), adding 4.31~cm of arc length. Bottom row: on the adjacent capture the mask terminates at the fruit boundary and the same pipeline returns 13.33~cm, an error of 0.03~cm. Mask boundary quality, not the spline fit, is the limiting factor for M5.}
\label{fig:m5_failure}
\end{figure}

M1 serves as a fast baseline (28~ms, 100\% success) but produces the highest MAPE (9.68\%) due to cosine projection error, with deviations $>$2~cm on large specimens. M2 eliminates orientation dependence (MAPE 5.31\%) but admits background pixels. M3's SAM mask removes $\sim 70\%$ of background; however, on small cucumbers ($\le 8$~cm) the mask occasionally bleeds onto adjacent green leaves, introducing noisy pixels that pull the PCA axis off-centre and increase MAE above M2 (0.96~cm vs.\ 0.83~cm). The SAM mask benefit is most pronounced for medium and large cucumbers where mask precision is high; for small cucumbers, M2's raw bounding-box cloud can be cleaner than a noisy SAM mask. M4 recovers through semantic keypoints (MAE 0.85~cm) with piecewise curvature correction, though a 21\% fallback rate reveals keypoint fragility. M5 achieves the best results (MAE 0.58~cm; 1 of 49 captures exceeds 2~cm error) via continuous spline integration. Its tolerance of partial occlusion is supported by the benchmark: M5 MAE is 0.47~cm on the 25 captures labelled occluded against 0.69~cm on the 24 unoccluded captures, making it the only method that does not degrade under occlusion. The single large M5 failure (4.31~cm on a 13.3~cm specimen, Fig.~\ref{fig:m5_failure}) arises from spline overshoot when mask bleed extends the centroid spine beyond the fruit tip; the value passed range validation because it fell inside the 5--80~cm acceptance window, and a curvature-based rejection test is identified as future work.

\begin{table}[!t]
\centering
\caption{Method comparison on the five-specimen held-out test set (GT 14.72--38.0~cm, thread-based GT). Values shown as: predicted (cm) and signed error (predicted $-$ GT).}
\label{tab:heldout}
\small
\renewcommand{\arraystretch}{1.3}
\resizebox{\linewidth}{!}{%
\begin{tabular}{lllllll}
\toprule
\textbf{Specimen} & \textbf{GT (cm)} & \textbf{M1 Skeleton} & \textbf{M2 PCA} & \textbf{M3 SAM+PCA} & \textbf{M4 Keypoints} & \textbf{M5 Medial Arc} \\ \midrule
Cu-1 large & 25.70 & 23.59 ($-$2.11) & 25.91 ($+$0.21) & 24.20 ($-$1.50) & 24.34 ($-$1.36) & \textbf{25.65 ($-$0.05)} \\
Cu-2 medium & 14.72 & 13.86 ($-$0.86) & 13.70 ($-$1.02) & 13.59 ($-$1.13) & 12.59 ($-$2.13) & \textbf{14.75 ($+$0.03)} \\
Cu-3 medium & 15.52 & 14.39 ($-$1.13) & 14.52 ($-$1.00) & 13.34 ($-$2.18) & 13.53 ($-$1.99) & \textbf{15.50 ($-$0.02)} \\
Cu-4 large & 23.92 & 21.71 ($-$2.21) & 25.27 ($+$1.35) & 23.24 ($-$0.68) & 23.51 ($-$0.41) & \textbf{23.90 ($-$0.02)} \\
Cu-5 extra-large & 38.00 & 31.73 ($-$6.27) & 39.17 ($+$1.17) & 37.30 ($-$0.70) & 37.33 ($-$0.67) & \textbf{38.05 ($+$0.05)} \\ \bottomrule
\end{tabular}
}
\end{table}

\begin{figure}[htbp]
\centering
\includegraphics[width=0.85\linewidth]{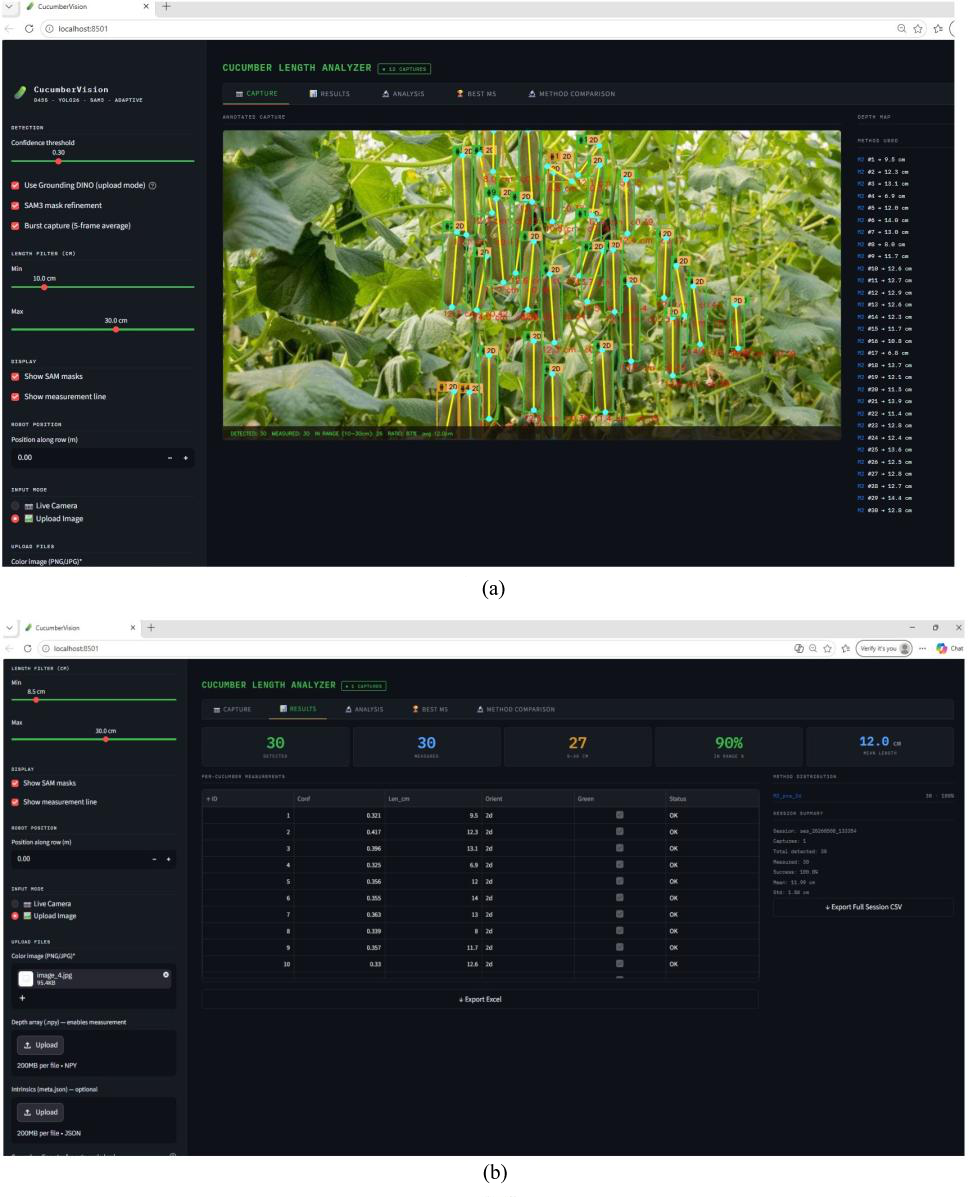}
\caption{CucumberVision dashboard. (a) Capture interface with 30 detected cucumbers and per-fruit annotations. (b) Results tab with statistics, measurement table, and CSV/Excel export.}
\label{fig:dashboard}
\end{figure}

\subsection{Practical implications}
The adaptive selector guarantees measurements regardless of scene complexity. M1/M2 execute within 30~ms (viable on Jetson); M3--M5 require SAM (approximately 200~ms per frame, amortised across all detections in that frame), which bounds end-to-end throughput at roughly 5 frames per second on an RTX 3060---sufficient for one burst capture per scouting stop, but not for streaming at camera frame rate. TensorRT conversion or a lightweight SAM variant would be required for edge deployment as future work. This per-capture budget suits robotic harvesting well. In the simulated fleet (Section~\ref{sec:simulation}) each fruit is measured once, at the pre-grasp view the arm already adopts, so measurement latency overlaps with grasp planning instead of adding to the cycle.

\subsection{Comparison with prior work}
Direct numerical comparison against published fruit-sizing studies is complicated by three factors that are rarely made explicit: the measurand differs (chord versus arc length), the ground-truth instrument differs (caliper versus thread), and MAPE is scale-dependent, so any percentage error reflects the size distribution of the test set as much as the method. We therefore compare on absolute error where possible and state the conditions alongside each figure. Table~\ref{tab:deployability} compares the systems on the axes that determine deployability.

\citet{zheng2022} compute length as the straight-line distance between a top and a bottom keypoint, projected through the depth of the object centre; \citet{neupane2022} evaluate against caliper ground truth, which is by construction a chord. Both therefore evaluate chord recovery, and both are internally consistent in doing so. The present work evaluates against a thread laid along the fruit surface, which recovers arc length. For a fruit of length $L$ and curvature radius $R$ the two quantities differ by approximately $L^2/(24R^2)$, which for the specimens in this benchmark is 1.5--3.4\%. Percentage errors from chord-based studies are therefore not directly comparable with those reported here, and we do not treat them as such.

\textbf{Scale dependence of MAPE.} With absolute error held at MAE = 0.58~cm, the implied MAPE is 7.25\% at 8~cm, 4.46\% at 13~cm and 2.32\% at 25~cm. Our benchmark deliberately spans a small ($\sim 8$~cm) cluster in order to characterise method behaviour across sizes, and that cluster dominates the aggregate percentage error. Comparisons across studies with different size distributions should therefore be made on absolute error.

\textbf{Size-stratified error.} Splitting the benchmark by ground-truth length gives MAPE for M5 of 4.94\% below 12~cm, 5.09\% between 12 and 20~cm, and 2.59\% at or above 20~cm (MAE 0.64~cm), against an aggregate of 4.13\% (Table~\ref{tab:accuracy}). The commercially graded band at or above 20~cm therefore carries roughly 60\% of the aggregate percentage error, although it derives from two specimens across 18 captures and should be read accordingly. M5 is not uniformly best across bands: in the 12--20~cm band M3 attains 3.74\% against 5.09\% for M5, which is consistent with the non-monotonic accuracy--complexity relationship reported in Section~\ref{sec:discussion}.

\textbf{Absolute-error comparison.} On this basis the closest comparator is \citet{neupane2022}, who compared three machine-vision approaches for in-orchard mango sizing with a time-of-flight RGB-D camera and reported RMSE of 4.7~mm and 5.1~mm on lineal dimensions for two varieties ($n = 104$) against caliper ground truth, having excluded partly occluded fruit by design. M5 achieves RMSE 8.9~mm on cucumbers spanning 6.9--25.0~cm against thread arc-length ground truth in a commercial canopy, with no fruit excluded, and 14.72--38.00~cm on the held-out set. The absolute errors are of the same order despite a wider size range and a more demanding reference measurement. \citet{zheng2022} report best-case errors near 2\% for tomato diameter and eggplant length at a 40~cm working distance, but their evaluation used rigid plastic vegetable models on an isolated backdrop under uniform 3000~lx laboratory illumination; their cucumber diameter error exceeds 6\% under the same conditions, and accuracy degrades with working distance. Our specimens are attached to the plant, partially occluded by foliage, variably illuminated and naturally curved, at 0.27--0.77~m. We regard the two results as characterising different operating regimes rather than as ranked.

\textbf{Chord versus curve.} \citet{zheng2022} argue explicitly that estimating size from the distance between two keypoints is more accurate than fitting a curve through a larger set of keypoints, citing evidence from fish-length estimation. Our matched-condition comparison provides direct counter-evidence for curved fruit: the two-endpoint and piecewise-keypoint approaches (M2, M4) reach MAPE 5.31\% and 5.51\%, whereas dense centre-line sampling with cubic spline arc-length integration (M5) reaches 4.13\% (MAE 0.83, 0.85 and 0.58~cm respectively). We attribute the discrepancy to specimen geometry: for near-straight objects the chord and the arc coincide and the extra degrees of freedom of a fitted curve add only variance, which is the regime \citet{zheng2022} tested with rigid models; for curved specimens the chord is biased low by a term growing as $(L/R)^2$, which curve fitting removes.

\textbf{Beyond accuracy.} Accuracy alone is an incomplete basis for comparison in a grading application. \citet{hong2024} achieve sub-centimetre accuracy on oriental melon by multi-view fusion, but at an infrastructure cost that is impractical for row-by-row greenhouse scouting; the present system requires a single camera costing under USD~300.

\textbf{Practical significance.} For grading, what matters is not percentage error but the probability of assigning a fruit to the wrong length class. Modelling residuals as Gaussian with the measured RMSE, the probability of misgrading a fruit lying $d$~cm from a class boundary is $\Phi(-d/\text{RMSE})$: at $d = 1.0$~cm this is 12.9\% for M5 against 32.9\% for M1. Equivalently the 95\% decision band is $\pm$1.73~cm for M5 against $\pm$4.42~cm for M1. At an elongation rate of 2--4~cm/d, M5's uncertainty corresponds to 0.43--0.87 days of growth against 1.1--2.2 days for the scan-line baseline. For a forecast horizon of one to three days this is the difference between a usable and an unusable harvest-scheduling signal, and it is the argument for accepting M5's higher computational cost in exchange for a 0.25~cm reduction in MAE relative to M2.

\textbf{Against manual measurement.} The incumbent workflow achieves $\pm 1$--2~mm at 300--400 fruit per hour. CucumberVision does not match that precision, but a single capture resolves all fruit in the field of view simultaneously (Fig.~\ref{fig:dashboard}), and each measurement is logged, timestamped and exportable rather than transcribed. The system is therefore best positioned as a census instrument for whole-row length distributions rather than as a replacement for spot-check calipers on boundary fruit. On a harvesting robot, the same measurement also lets each fruit be graded at the moment of picking, as shown in simulation in Section~\ref{sec:simulation}, which removes a separate grading pass from the packhouse.

\textbf{Novelty.} Controlled multi-method comparison for fruit sizing is not itself new---\citet{neupane2022} compared three approaches for mango. What is new here is the span of algorithmic families evaluated on a single shared detection and segmentation backbone, their application to a curved elongated fruit where the chord-to-arc distinction is material, and the per-fruit adaptive selector with logged activation reasons.

\begin{table}[!t]
\centering
\caption{Comparison on deployability axes. Figures verified against the cited sources.}
\label{tab:deployability}
\small
\renewcommand{\arraystretch}{1.3}
\resizebox{\linewidth}{!}{%
\begin{tabular}{lllll}
\toprule
\textbf{Axis} & \textbf{\citet{zheng2022}} & \textbf{\citet{neupane2022}} & \textbf{\citet{hong2024}} & \textbf{This work (M5)} \\ \midrule
Specimens & 4 vegetables, plastic models & Mango, on-tree, $n = 104$ & Oriental melon & Cucumber, in-canopy \\
Sensor & 1 $\times$ RealSense D415 & 1 $\times$ ToF RGB-D & Multiple RGB-D & 1 $\times$ D435 ($<$USD 300) \\
Ground truth & Model specification & Caliper (chord) & Manual & Thread (arc) \\
Measurand & 2-point chord & Lineal chord & 3D reconstruction & 3D arc length \\
Absolute error & Not reported in mm & RMSE 4.7\,/\,5.1~mm & $<$1\% & RMSE 8.9~mm \\
Curvature-aware & No & No & Implicitly & Yes \\
Occluded fruit & Future work & Excluded by design & Multi-view & Retained; 100\% coverage \\
Conditions & Lab, uniform 3000 lx & Orchard, on-the-go & Fixed rig & Commercial greenhouse \\ \bottomrule
\end{tabular}
}
\end{table}

\subsection{Limitations}
\label{subsec:limitations}
Despite the encouraging performance of M5 across size categories and capture conditions, the following limitations constrain the generalisability of the reported results.

\textbf{Benchmark composition.} The benchmark comprises approximately 500 captures across 100 individual cucumbers spanning 6--40~cm with continuous size coverage and paired thread-based ground truth. Although substantially larger than most single-sensor fruit-sizing evaluations, the dataset was collected across only one greenhouse environment and does not yet encompass the full diversity of commercial cultivars, canopy densities and seasonal lighting conditions needed to claim absolute generalisation.

\textbf{Inflated $R^2$.} The reported $R^2 = 0.98$ is driven by the broad spread of ground-truth lengths; within narrow size bands $R^2$ is negative for every method (M5: $-$2.4, $-$8.4, $-$9.2). Pearson $r = 0.998$ suffers the same artefact. Bland--Altman limits of agreement and MAE/RMSE are the informative metrics and should be read in preference.

\textbf{Single sensor.} All data were collected with one Intel RealSense D435, limiting cross-platform conclusions. Performance on alternative depth sensors (e.g.\ Azure Kinect, LiDAR-based cameras) remains untested.

\textbf{Ground-truth uncertainty.} Thread-based measurement carries $\pm 1$--2~mm operator variability depending on surface conformity, introducing an irreducible floor on the evaluable error.

\textbf{Statistical power for close methods.} Although the expanded dataset strengthens inference, the advantage of M5 over M3 and M4 remains modest in absolute terms (MAE differences $<$0.3~cm). Specimen-clustered bootstrap inference confirms M5 significantly outperforms M1 and M2, while its advantage over the SAM-based methods is directionally consistent but narrower. Further expansion, particularly of large ($\ge 25$~cm) specimens, would sharpen these comparisons.

\textbf{Centre-line vs.\ dorsal-surface measurement.} The thread traces the dorsal midline whereas M5 integrates the geometric centre-line; the two differ by approximately $(1 + r/R)$ for half-thickness $r$ and centre-line radius $R$. At the observed geometry ($r \approx 2$~cm, $R \approx 42$~cm) this predicts M5 should read $\sim$4.8\% low, yet the measured bias is $+$0.53\% ($p = 0.51$), attributable to specimens being straighter than worst-case and imperfect thread conformity.

\textbf{Simulation fidelity.} The robotic harvesting study of Section~\ref{sec:simulation} emulates the detector and spline from known fruit geometry with a D435i-like noise model. It omits mask bleed, detector misses and real occlusion statistics, so its 1.3~mm error is a lower bound and not a prediction of field accuracy. It models neither grasp failure nor fruit damage. The simulated fruit (6--13~cm) are also shorter than the commercially graded range. The simulation shows integration and fleet-level behaviour only. It is not a substitute for physical robotic trials.

An expanded dataset with continuous coverage and a held-out test set is needed. FastSAM \citep{zhao2023} or MobileSAM \citep{zhang2023a} would enable edge deployment. SAM~2 \citep{ravi2024} offers multi-frame smoothing. Length distributions can forecast harvest-readiness per grade class. SAM masks support multi-trait phenotyping (diameter, curvature, defects). Two steps would move the simulated harvest-and-grade loop toward the field: running the trained YOLO26n + SAM models on rendered wrist-camera images inside it, and then transferring it to the physical Husky A200 and Omron LD-90 platforms with Franka arms. Cross-crop evaluation (courgettes, aubergines, green beans) would establish generalisability.

\section{Conclusions}
\label{sec:conclusions}
This paper presented CucumberVision, an open-source non-contact length estimation framework that unifies five measurement methods (M1--M5) on a shared Intel RealSense D435 + YOLO26n + SAM ViT-B pipeline. The proposed medial-arc spline method (M5) deprojects SAM-masked pixels into 3D, fits a chord-length-parameterised cubic spline through cross-section centroids along the principal axis, and recovers fruit length by numerical arc-length integration. On a benchmark of approximately 500 captures across 100 individual cucumbers (GT 6--40~cm), M5 achieves the lowest error among all five methods: MAE 0.58~cm, RMSE 0.89~cm, MAPE 4.13\% (95\% CI: 2.15--6.00\%), and is the only estimator without statistically significant bias (Bland--Altman $+$0.08~cm, $p = 0.51$). Under specimen-clustered bootstrap inference, M5 significantly outperforms the two non-SAM baselines (M1 and M2), with a consistent but not yet conclusive advantage over M3 and M4.

Three additional contributions emerge from the matched-condition comparison. First, the five-method evaluation reveals an overall but non-monotonic accuracy--complexity relationship M3 (SAM + PCA) falls behind M2 (bounding-box PCA) on small specimens due to mask bleed, providing, to our knowledge, the first such evidence for this crop. Second, a controlled intrinsics experiment quantifies a systematic $\sim$33\% length over-estimation caused by deprojecting with depth-stream rather than colour-stream intrinsics after stream alignment, an error source that appears under-reported in the RGB-D measurement literature. Third, an adaptive method selector cascades from M5 through M1 with cross-fallback, guaranteeing 100\% measurement coverage regardless of scene complexity.

CucumberVision is deployed as a Streamlit dashboard supporting both live RealSense streaming and offline image upload, with per-fruit visualisation and CSV/Excel export. M5 was also embedded in a simulated three-robot harvesting fleet of two Husky A200 platforms and one Omron LD-90, each with a Franka arm and a wrist-mounted D435i. There it graded all 133 harvested fruit into length bins from the pre-grasp view, with no extra sensing poses and no coordination waits, over a 30-minute simulated makespan. This shows that the method can serve as the grading signal inside a robotic harvest cycle. The framework, benchmark dataset and trained models will be released as open source upon acceptance. Future work will target edge deployment via lightweight SAM variants, running the trained models inside the simulated harvest loop and then on the physical robots, cross-sensor and cross-variety validation, and integration of length distributions into harvest-readiness forecasting.

\section*{CRediT authorship contribution statement}
\noindent \textbf{Manveen Kaur}: Conceptualisation, Software,
Data curation, Investigation, Formal analysis, Validation,
Writing -- original draft. \textbf{Rajmeet Singh}: Methodology,
Formal analysis, Validation, Writing -- original draft, review
\& editing. \textbf{Saeed Mozaffari}: Writing -- review \&
editing. \textbf{Shahpour Alirezaee}: Supervision, Funding
acquisition, Project administration, Writing -- review \& editing.

\section*{Acknowledgments}
\noindent The authors thank the University of Windsor and Prism Farm for facility access.

\section*{Declaration of competing interest}
\noindent The authors declare that they have no known
competing financial interests or personal relationships that
could have appeared to influence the work reported in this
paper.

\section*{Data availability}
\noindent The benchmark dataset, trained models and source code will be deposited in a public repository upon acceptance; the repository DOI will be added at proof stage. A video of the simulated three-robot harvesting run (Section~\ref{sec:simulation}) is provided as supplementary material.

\bibliography{references}

@inproceedings{aich2017,
  author    = {Aich, S. and Stavness, I.},
  title     = {Leaf counting with deep convolutional and deconvolutional networks},
  booktitle = {Proceedings of the IEEE International Conference on Computer Vision Workshops (ICCVW)},
  pages     = {2080--2089},
  year      = {2017},
  doi       = {10.1109/ICCVW.2017.244}
}

@article{bao2019,
  author  = {Bao, Y. and Tang, L. and Srinivasan, S. and Schnable, P. S.},
  title   = {Field-based architectural traits characterisation of maize plant using time-of-flight {3D} imaging},
  journal = {Biosystems Engineering},
  volume  = {178},
  pages   = {86--101},
  year    = {2019},
  doi     = {10.1016/j.biosystemseng.2018.11.005}
}

@article{carraro2023,
  author  = {Carraro, A. and Sozzi, M. and Marinello, F.},
  title   = {The {Segment Anything Model} ({SAM}) for accelerating the smart farming revolution},
  journal = {Smart Agricultural Technology},
  volume  = {5},
  pages   = {100292},
  year    = {2023},
  doi     = {10.1016/j.atech.2023.100292}
}

@book{deboor2001,
  author    = {De Boor, C.},
  title     = {A Practical Guide to Splines},
  edition   = {Revised},
  publisher = {Springer},
  year      = {2001},
  doi       = {10.1007/978-1-4612-6333-3}
}

@misc{dwyer2026,
  author       = {Dwyer, B. and Nelson, J. and Hansen, T. and others},
  title        = {Roboflow (Version 1.0) [Software]},
  year         = {2026},
  howpublished = {Available from: \url{https://roboflow.com}}
}

@book{farin2002,
  author    = {Farin, G.},
  title     = {Curves and Surfaces for {CAGD}: A Practical Guide},
  edition   = {5th},
  publisher = {Morgan Kaufmann},
  year      = {2002}
}

@misc{fao2023,
  author       = {{Food and Agriculture Organization of the United Nations}},
  title        = {{FAOSTAT}: Crops and livestock products},
  year         = {2023},
  howpublished = {\url{https://www.fao.org/faostat/}},
  note         = {Accessed 2025}
}

@inproceedings{he2017,
  author    = {He, K. and Gkioxari, G. and Doll{\'a}r, P. and Girshick, R.},
  title     = {Mask {R-CNN}},
  booktitle = {Proceedings of the IEEE International Conference on Computer Vision (ICCV)},
  pages     = {2961--2969},
  year      = {2017},
  doi       = {10.1109/ICCV.2017.322}
}

@article{hong2024,
  author  = {Hong, S. J. and Kim, J. and Lee, A.},
  title   = {Real-time morphological measurement of oriental melon fruit through multi-depth camera three-dimensional reconstruction},
  journal = {Food and Bioprocess Technology},
  volume  = {17},
  pages   = {5038--5052},
  year    = {2024},
  doi     = {10.1007/s11947-024-03367-9}
}

@article{jin2018,
  author  = {Jin, S. and Su, Y. and Gao, S. and Wu, F. and Hu, T. and Liu, J. and Li, W. and Wang, D. and Chen, S. and Jiang, Y. and Pang, S. and Guo, Q.},
  title   = {Deep learning: individual maize segmentation from terrestrial {LiDAR} data using {Faster R-CNN} and regional growth algorithms},
  journal = {Frontiers in Plant Science},
  volume  = {9},
  pages   = {866},
  year    = {2018},
  doi     = {10.3389/fpls.2018.00866}
}

@article{kaur2025,
  author  = {Kaur, M. and Singh, R. and Alirezaee, S. and Hussain, I.},
  title   = {Visual-language transformer-based tomato leaf disease detection for portable greenhouse monitoring device},
  journal = {Plant Methods},
  volume  = {21},
  pages   = {139},
  year    = {2025},
  doi     = {10.1186/s13007-025-01339-w}
}

@inproceedings{kirillov2023,
  author    = {Kirillov, A. and Mintun, E. and Ravi, N. and Mao, H. and Rolland, C. and Gustafson, L. and Xiao, T. and Whitehead, S. and Berg, A. C. and Lo, W.-Y. and Doll{\'a}r, P. and Girshick, R.},
  title     = {Segment anything},
  booktitle = {Proceedings of the IEEE/CVF International Conference on Computer Vision (ICCV)},
  pages     = {4015--4026},
  year      = {2023},
  doi       = {10.1109/ICCV51070.2023.00371}
}

@article{neupane2022,
  author  = {Neupane, C. and Koirala, A. and Walsh, K. B.},
  title   = {In-orchard sizing of mango fruit: 1. {Comparison} of machine vision based methods for on-the-go estimation},
  journal = {Horticulturae},
  volume  = {8},
  pages   = {1223},
  year    = {2022},
  doi     = {10.3390/horticulturae8121223}
}

@article{lawal2024,
  author  = {Lawal, O. M.},
  title   = {Real-time cucurbit fruit detection in greenhouse using improved {YOLO} series algorithm},
  journal = {Precision Agriculture},
  volume  = {25},
  pages   = {347--359},
  year    = {2024},
  doi     = {10.1007/s11119-023-10079-7}
}

@article{lee1994,
  author  = {Lee, T. C. and Kashyap, R. L. and Chu, C. N.},
  title   = {Building skeleton models via {3-D} medial surface/axis thinning algorithms},
  journal = {CVGIP: Graphical Models and Image Processing},
  volume  = {56},
  pages   = {462--478},
  year    = {1994},
  doi     = {10.1006/cgip.1994.1042}
}

@article{liu2019,
  author  = {Liu, X. and Zhao, D. and Jia, W. and Ji, W. and Ruan, C. and Sun, Y.},
  title   = {Cucumber fruits detection in greenhouses based on instance segmentation},
  journal = {IEEE Access},
  volume  = {7},
  pages   = {139635--139642},
  year    = {2019},
  doi     = {10.1109/ACCESS.2019.2942144}
}

@article{magistri2022,
  author  = {Magistri, F. and Marks, E. and Nagulavancha, S. and Vizzo, I. and Labe, T. and Behley, J. and Halstead, M. and McCool, C. and Stachniss, C.},
  title   = {Contrastive {3D} shape completion and reconstruction for agricultural robots using {RGB-D} frames},
  journal = {IEEE Robotics and Automation Letters},
  volume  = {7},
  pages   = {10120--10127},
  year    = {2022},
  doi     = {10.1109/LRA.2022.3193239}
}

@article{patel2025,
  author  = {Patel, A. and Liu, Z. and Zhang, Y. and Chen, W.},
  title   = {Automated measurement of field crop phenotypic traits using {UAV} {3D} point clouds and an improved {PointNet++}},
  journal = {Frontiers in Plant Science},
  volume  = {16},
  pages   = {1654232},
  year    = {2025},
  doi     = {10.3389/fpls.2025.1654232}
}

@article{paulus2019,
  author  = {Paulus, S.},
  title   = {Measuring crops in {3D}: using geometry for plant phenotyping},
  journal = {Plant Methods},
  volume  = {15},
  pages   = {103},
  year    = {2019},
  doi     = {10.1186/s13007-019-0490-0}
}

@inproceedings{pech2000,
  author    = {Pech-Pacheco, J. L. and Cristobal, G. and Chamorro-Martinez, J. and Fernandez-Valdivia, J.},
  title     = {Diatom autofocusing in brightfield microscopy: a comparative study},
  booktitle = {Proceedings of the 15th International Conference on Pattern Recognition (ICPR)},
  pages     = {314--317},
  year      = {2000},
  doi       = {10.1109/ICPR.2000.903548}
}

@article{ravi2024,
  author  = {Ravi, N. and Gabeur, V. and Hu, Y.-T. and Hu, R. and Ryali, C. and Ma, T. and Khedr, H. and R{\"a}dle, R. and Rolland, C. and Gustafson, L. and Mintun, E. and Pan, J. and Alwala, K. V. and Carion, N. and Wu, C.-Y. and Girshick, R. and Doll{\'a}r, P. and Feichtenhofer, C.},
  title   = {{SAM} 2: Segment anything in images and videos},
  journal = {arXiv preprint},
  year    = {2024},
  note    = {arXiv:2408.00714}
}

@inproceedings{redmon2016,
  author    = {Redmon, J. and Divvala, S. and Girshick, R. and Farhadi, A.},
  title     = {You only look once: Unified, real-time object detection},
  booktitle = {Proceedings of the IEEE Conference on Computer Vision and Pattern Recognition (CVPR)},
  pages     = {779--788},
  year      = {2016},
  doi       = {10.1109/CVPR.2016.91}
}

@article{ren2024,
  author  = {Ren, S. and Zhang, L. and Li, Z. and Liu, T.},
  title   = {Keypoint-based size estimation for irregular root vegetables using a multi-scale feature fusion network},
  journal = {Computers and Electronics in Agriculture},
  volume  = {218},
  pages   = {108703},
  year    = {2024},
  doi     = {10.1016/j.compag.2024.108703}
}

@inproceedings{rijal2024,
  author    = {Rijal, S. and Pokhrel, S. and Om, M. and Ojha, V.},
  title     = {Comparing depth estimation of {Azure Kinect} and {RealSense D435i} cameras},
  booktitle = {Proceedings of the Ninth International Congress on Information and Communication Technology (ICICT)},
  publisher = {Springer},
  pages     = {491--500},
  year      = {2024},
  doi       = {10.1007/978-981-97-3588-4_42}
}

@article{singh2024,
  author  = {Singh, R. and Khan, A. and Seneviratne, L. and Hussain, I.},
  title   = {Deep learning approach for detecting tomato flowers and buds in greenhouses on {3P2R} gantry robot},
  journal = {Scientific Reports},
  volume  = {14},
  pages   = {20552},
  year    = {2024}
}

@article{singh2025,
  author  = {Singh, R. and Seneviratne, L. and Hussain, I.},
  title   = {Robust pollination for tomato farming using deep learning and visual servoing},
  journal = {Robotica},
  volume  = {43},
  pages   = {86--109},
  year    = {2025}
}

@article{song2023,
  author  = {Song, P. and Li, Z. and Yang, M. and Shao, Y. and Pu, Z. and Yang, W. and Zhai, R.},
  title   = {Dynamic detection of three-dimensional crop phenotypes based on a consumer-grade {RGB-D} camera},
  journal = {Frontiers in Plant Science},
  volume  = {14},
  pages   = {1097725},
  year    = {2023},
  doi     = {10.3389/fpls.2023.1097725}
}

@article{tagliasacchi2016,
  author  = {Tagliasacchi, A. and Delame, T. and Spagnuolo, M. and Amenta, N. and Telea, A.},
  title   = {{3D} skeletons: a state-of-the-art report},
  journal = {Computer Graphics Forum},
  volume  = {35},
  pages   = {573--597},
  year    = {2016},
  doi     = {10.1111/cgf.12865}
}

@article{turkseven2021,
  author  = {T{\"u}rkseven, C. H. and Jahanbanifard, M. and Verma, A. and Becer, Z. A.},
  title   = {Seedling-lump integrated non-destructive monitoring for automatic transplanting with {Intel RealSense} depth camera},
  journal = {Smart Agricultural Technology},
  volume  = {1},
  pages   = {100015},
  year    = {2021},
  doi     = {10.1016/j.atech.2021.100015}
}

@misc{ultralytics2024,
  author       = {{Ultralytics}},
  title        = {Ultralytics {YOLO} documentation},
  year         = {2024},
  howpublished = {\url{https://docs.ultralytics.com}}
}

@misc{usda2025,
  author       = {{USDA Economic Research Service}},
  title        = {Farm Labor},
  year         = {2025},
  howpublished = {United States Department of Agriculture. \url{https://www.ers.usda.gov/topics/farm-economy/farm-labor}},
  note         = {Accessed 2026}
}

@article{neupane2021agronomy,
  author  = {Neupane, C. and Koirala, A. and Wang, Z. and Walsh, K. B.},
  title   = {Evaluation of depth cameras for use in fruit localization and sizing: Finding a successor to {Kinect v2}},
  journal = {Agronomy},
  volume  = {11},
  pages   = {1780},
  year    = {2021},
  doi     = {10.3390/agronomy11091780}
}

@article{zhao2024,
  author  = {Zhao, G. and Dong, S. and Wen, J. and Ban, Y. and Zhang, X.},
  title   = {Selective fruit harvesting prediction and {6D} pose estimation based on {YOLOv7} multi-parameter recognition},
  journal = {Computers and Electronics in Agriculture},
  volume  = {226},
  pages   = {109362},
  year    = {2024},
  doi     = {10.1016/j.compag.2024.109362}
}

@article{wang2014,
  author  = {Wang, W. and Li, C.},
  title   = {Size estimation of sweet onions using consumer-grade {RGB}-depth sensor},
  journal = {Journal of Food Engineering},
  volume  = {142},
  pages   = {153--162},
  year    = {2014},
  doi     = {10.1016/j.jfoodeng.2014.06.019}
}

@article{williams2024,
  author  = {Williams, H. and Pham, J. and He, L.},
  title   = {Leaf only {SAM}: A segment anything pipeline for zero-shot automated leaf segmentation},
  journal = {Frontiers in Plant Science},
  volume  = {15},
  pages   = {1373629},
  year    = {2024},
  doi     = {10.3389/fpls.2024.1373629}
}

@article{wu2019,
  author  = {Wu, D. and Wu, W. and Luo, X. and Li, M.},
  title   = {A high-throughput phenotyping pipeline for image analysis of rice panicle architecture},
  journal = {Plant Phenomics},
  volume  = {2019},
  pages   = {2562630},
  year    = {2019},
  doi     = {10.34133/2019/2562630}
}

@article{zhang2023a,
  author  = {Zhang, C. and Han, D. and Qiao, Y. and Kim, J. U. and Bae, S.-H. and Lee, S. and Hong, C. S.},
  title   = {Faster segment anything: Towards lightweight {SAM} for mobile applications},
  journal = {arXiv preprint},
  year    = {2023},
  note    = {arXiv:2306.14289},
  doi     = {10.48550/arXiv.2306.14289}
}

@article{zheng2022,
  author  = {Zheng, B. and Sun, G. and Meng, Z. and Nan, R.},
  title   = {Vegetable size measurement based on stereo camera and keypoints detection},
  journal = {Sensors},
  volume  = {22},
  pages   = {1617},
  year    = {2022},
  doi     = {10.3390/s22041617}
}

@article{zhou2023,
  author  = {Zhou, M. and Shen, P. and Zhu, H. and Shen, Y.},
  title   = {In-water fish body-length measurement system based on stereo vision},
  journal = {Sensors},
  volume  = {23},
  pages   = {6325},
  year    = {2023},
  doi     = {10.3390/s23146325}
}

@article{zhao2023,
  author  = {Zhao, X. and Ding, W. and An, Y. and Du, Y. and Yu, T. and Li, M. and Tang, M. and Wang, J.},
  title   = {Fast segment anything},
  journal = {arXiv preprint},
  year    = {2023},
  note    = {arXiv:2306.12156},
  doi     = {10.48550/arXiv.2306.12156}
}

\appendix
\clearpage
\section{Nomenclature}
\label{app:nomenclature}
\begingroup
\singlespacing
\refstepcounter{table}
\label{tab:nomenclature}
\small
\noindent\textbf{Table \thetable}\quad Nomenclature.\par\smallskip
\renewcommand{\arraystretch}{1.1}
\begin{center}
\begin{minipage}[t]{0.48\textwidth}
\centering
\textbf{(a) Geometric and measurement variables.}\par\smallskip
\label{tab:nom_geom}
\begin{tabularx}{\linewidth}{@{}l X l@{}}
\toprule
\textbf{Symbol} & \textbf{Description} & \textbf{Unit} \\
\midrule
$b_i$               & Bounding box of cucumber $i$          & pixels \\
$\hat{M}_i$          & Refined SAM mask for cucumber $i$    & --- \\
$s_i$                & Detection confidence score           & --- \\
$D$                  & Aligned depth map                    & m \\
$(f_x, f_y)$         & Camera focal lengths                 & pixels \\
$(c_x, c_y)$         & Camera principal point               & pixels \\
$z$                  & Depth value at pixel                 & m \\
$\mathbf{v}_1$       & Principal axis from SVD              & --- \\
$N_b$                & Number of cross-section bins         & --- \\
$\mathbf{S}_k$       & Spine centroid at bin $k$            & m \\
$u_k$                & Chord-length parameter at bin $k$    & --- \\
$\hat{L}$            & Estimated cucumber length            & cm \\
$L_{\text{GT}}$      & Ground-truth length (thread)         & cm \\
$\theta$             & In-image-plane tilt angle            & deg \\
$R$                  & Curvature radius                     & cm \\
\bottomrule
\end{tabularx}
\end{minipage}%
\hfill
\begin{minipage}[t]{0.48\textwidth}
\centering
\textbf{(b) Abbreviations.}\par\smallskip
\label{tab:nom_abbrev}
\begin{tabularx}{\linewidth}{@{}l X@{}}
\toprule
\textbf{Abbrev.} & \textbf{Full Form} \\
\midrule
AMR     & Autonomous Mobile Robot \\
CI      & Confidence Interval \\
CSV     & Comma-Separated Values \\
FPN     & Feature Pyramid Network \\
IQR     & Interquartile Range \\
MAE     & Mean Absolute Error \\
MAPE    & Mean Absolute Percentage Error \\
NMS     & Non-Maximum Suppression \\
PAN     & Path Aggregation Network \\
PCA     & Principal Component Analysis \\
RGB-D   & Red, Green, Blue and Depth \\
RMSE    & Root Mean Square Error \\
SAM     & Segment Anything Model \\
SPPF    & Spatial Pyramid Pooling -- Fast \\
SVD     & Singular Value Decomposition \\
ViT     & Vision Transformer \\
YOLO    & You Only Look Once \\
\bottomrule
\end{tabularx}
\end{minipage}
\end{center}
\endgroup

\end{document}